\documentclass[onecolumn]{viocean}
\pdfoutput=1

\usepackage[toc,page,header]{appendix}

\usepackage{natbib}
\usepackage{latexsym}

\usepackage{url}
\usepackage{amssymb}
\usepackage[utf8]{inputenc}
\usepackage{microtype}
\usepackage{booktabs}
\usepackage{pifont} 
\usepackage{multirow}
\usepackage{makecell}
\usepackage{paralist}
\usepackage{xspace}
\usepackage{color}
\usepackage{xcolor}
\usepackage{colortbl}
\usepackage{adjustbox}
\usepackage{hyperref} 
\usepackage[edges]{forest}
\usepackage{tikz} 
\usepackage{caption}
\usepackage{amsfonts}

\hypersetup{
    colorlinks,
    linkcolor={blue!80!black},
    citecolor={blue!80!black},
}
\tikzset{
    root/.style =             {align=center, text width=1cm, rounded corners=3pt, line width=0.3mm, fill=gray!10, draw=gray!80, font=\small},
    demographic/.style =         {align=center, text width=1.8cm, rounded corners=3pt, line width=0.3mm, fill=blue!10, draw=blue!80, font=\footnotesize},
    demographic_work/.style =    {align=center, text width=10cm, rounded corners=3pt, line width=0.3mm, fill=blue!10, draw=blue!0, font=\footnotesize},
    character/.style =         {align=center, text width=1.8cm, rounded corners=3pt, line width=0.3mm, fill=red!10, draw=red!80, font=\footnotesize},
    character_work/.style =    {align=center, text width=10cm, rounded corners=3pt, line width=0.3mm, fill=red!10, draw=red!0, font=\footnotesize},
    personalization/.style =           {align=center, text width=1.8cm, rounded corners=3pt, line width=0.3mm, fill=cyan!10, draw=cyan!80, font=\footnotesize},
    personalization_work/.style =      {align=center, text width=10cm, rounded corners=3pt, line width=0.3mm, fill=cyan!10, draw=cyan!0, font=\footnotesize},
    risk/.style =         {align=center, text width=1.8cm, rounded corners=3pt, line width=0.3mm, fill=orange!10, draw=orange!80, font=\footnotesize},
    risk_work/.style =    {align=center, text width=10cm, rounded corners=3pt, line width=0.3mm, fill=orange!10, draw=orange!0, font=\footnotesize},
}

\newcommand{\eg}{\textit{e.g.}\xspace}

\usepackage{CJK}

\usepackage{amsmath,amsfonts,bm}

\def\eqref#1{equation~\ref{#1}}

\def\1{\bm{1}}

\def\vs{{\bm{s}}}

\DeclareMathAlphabet{\mathsfit}{\encodingdefault}{\sfdefault}{m}{sl}
\SetMathAlphabet{\mathsfit}{bold}{\encodingdefault}{\sfdefault}{bx}{n}

\usepackage{url}

\usepackage{amsmath}
\usepackage{amssymb}
\usepackage{bm,bbm}
\usepackage[makeroom]{cancel}
\usepackage{adjustbox}
\usepackage{xspace}
\usepackage{color}
\usepackage{xcolor}
\usepackage{colortbl}
\usepackage{subcaption}
\usepackage{multirow}
\usepackage{adjustbox}
\usepackage{booktabs}
\usepackage{makecell}
\usepackage{wrapfig}

\usepackage{tcolorbox}
\usepackage{amssymb}

\definecolor{iccvblue}{rgb}{0.21,0.49,0.74}
\definecolor{citecolor}{HTML}{2980b9}
\definecolor{linkcolor}{HTML}{c0392b}
\definecolor{backred}{RGB}{255, 190, 190}
\definecolor{backblue}{RGB}{210, 230, 250}
\definecolor{pink}{RGB}{240, 81, 121}
\definecolor{green}{RGB}{69, 189, 155}
\definecolor{yellow}{RGB}{253, 207, 110}
\definecolor{lightred}{RGB}{254, 129, 125}
\definecolor{lightblue}{RGB}{129, 184, 223}
\usepackage{utfsym} 
\newcommand{\heart}{$\;\!$\usym{2665}}
\newcolumntype{C}[1]{>{\centering\arraybackslash}m{#1}}
\makeatletter
  \newcommand\figcaption{\def\@captype{figure}\caption}
  \newcommand\tabcaption{\def\@captype{table}\caption}
\makeatother

\makeatletter
\DeclareRobustCommand\onedot{\futurelet\@let@token\mathbfv@onedotaux}
\def\mathbfv@onedotaux{\ifx\@let@token.\else.\null\fi\xspace}
\def\eg{\textit{e.g}\onedot}

 \def\vs{\textit{vs}\onedot}
\def\wrt{w.r.t\onedot}

\usepackage{relsize} 
\newcommand{\ourparser}{SymParser}
\newcommand{\ourrl}{SymHPR}
\newcommand{\ourvae}{SymVAE}

\usepackage{amsmath}
\usepackage{amssymb}
\usepackage{mathtools}
\usepackage{amsthm}
\usepackage{bm,bbm}
\usepackage{algorithm}
\usepackage{algpseudocode}

\title{Hierarchical Process Reward Models are Symbolic Vision Learners}

\author{%
    \textcolor{vioceanviolet}{Shan Zhang}\textsuperscript{1,5,$\dagger$,}\teams{\textcolor{vioceanviolet}{Researchers collaborating on the ViOcean initiative.}}\thanks{Core Contribution\quad $\dagger$Project Lead\quad $\ddagger$Corresponding Author} \quad
     \textcolor{vioceanviolet}{Aotian Chen}\textsuperscript{2,$\ast$}\teams{\textcolor{vioceanviolet}{Researchers collaborating on the ViOcean initiative.}}\quad
     {Kai Zou}\textsuperscript{3}\quad
    Jindong Gu\textsuperscript{4} \quad
    Yuan Xue\textsuperscript{2,$\ddagger$ }\quad 
    Anton van den Hengel\textsuperscript{1}\quad \\
    \affiliation[1]{Adelaide AIML\quad $^2$ Ohio State University\quad $^3$NetMind.ai\quad $^4$University of Oxford\quad $^5$Data61$\!${\color{red}$\heart$}CSIRO} 
    } 
\abstract{\begin{abstract}

Symbolic computer vision represents diagrams through explicit logical rules and structured representations, enabling interpretable understanding in machine vision. This requires fundamentally different learning paradigms from pixel-based visual models. Symbolic visual learners parse diagrams into geometric primitives—points, lines, and shapes—whereas pixel-based learners operate on textures and colors. 
We propose a novel self-supervised symbolic auto-encoder that encodes diagrams into structured primitives and their interrelationships within the latent space, and decodes them through our executable engine to reconstruct the input diagrams.
Central to this architecture is \ourrl{} (Symbolic Hierarchical Process Reward Modeling), which applies hierarchical step-level parsing rewards to enforce point-on-line, line-on-shape, and shape-on-relation consistency. Since vanilla reinforcement learning exhibits poor exploration in the policy space during diagram reconstruction; we thus introduce stabilization mechanisms to balance exploration and exploitation.

We fine-tune our symbolic encoder on downstream tasks, developing a neuro-symbolic system that integrates the reasoning capabilities of neural networks with the interpretability of symbolic models through reasoning-grounded \textit{visual rewards}. Evaluations across reconstruction, perception, and reasoning tasks demonstrate the effectiveness of our approach: achieving a 98.2\% reduction in MSE for geometric diagram reconstruction, surpassing GPT-4o by 0.6\% with a 7B model on chart reconstruction, and improving by +13\% on the MathGlance perception benchmark, and by +3\% on MathVerse and GeoQA reasoning benchmarks.

\end{abstract}}

\date{\today}  
\connection{\email{shan.zhang@adelaide.edu.au}, \email{Yuan.Xue@osumc.edu},
\email{anton.vandenhengel@adelaide.edu.au} }

\checkdata[Project Page]{\url{https://vi-ocean.github.io/projects/SymVAE}}

\abslogo{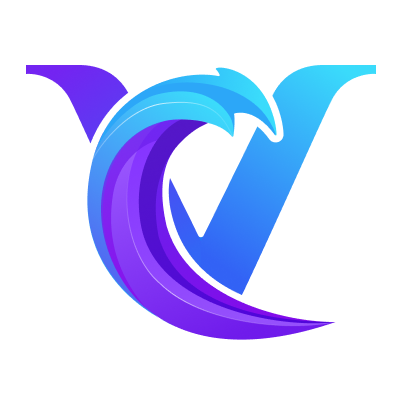} 

\begin{document}
\maketitle
\section{Introduction}


Across science, engineering, and education, diagrams are a universal medium for reasoning, design, and communication. From geometric proofs and scientific schematics to electrical circuits and data charts, diagrams encode structural relations that words alone cannot easily convey. They serve as a bridge between visual intuition and formal logic, allowing humans to externalize complex relationships in a compact and interpretable form—a process that embodies symbolic system, the parsing of explicit geometric primitives and their structural organization~\cite{mao2019neuro}.

\begin{wrapfigure}{r}{8.5cm}
    \centering
    \includegraphics[width=1\linewidth]{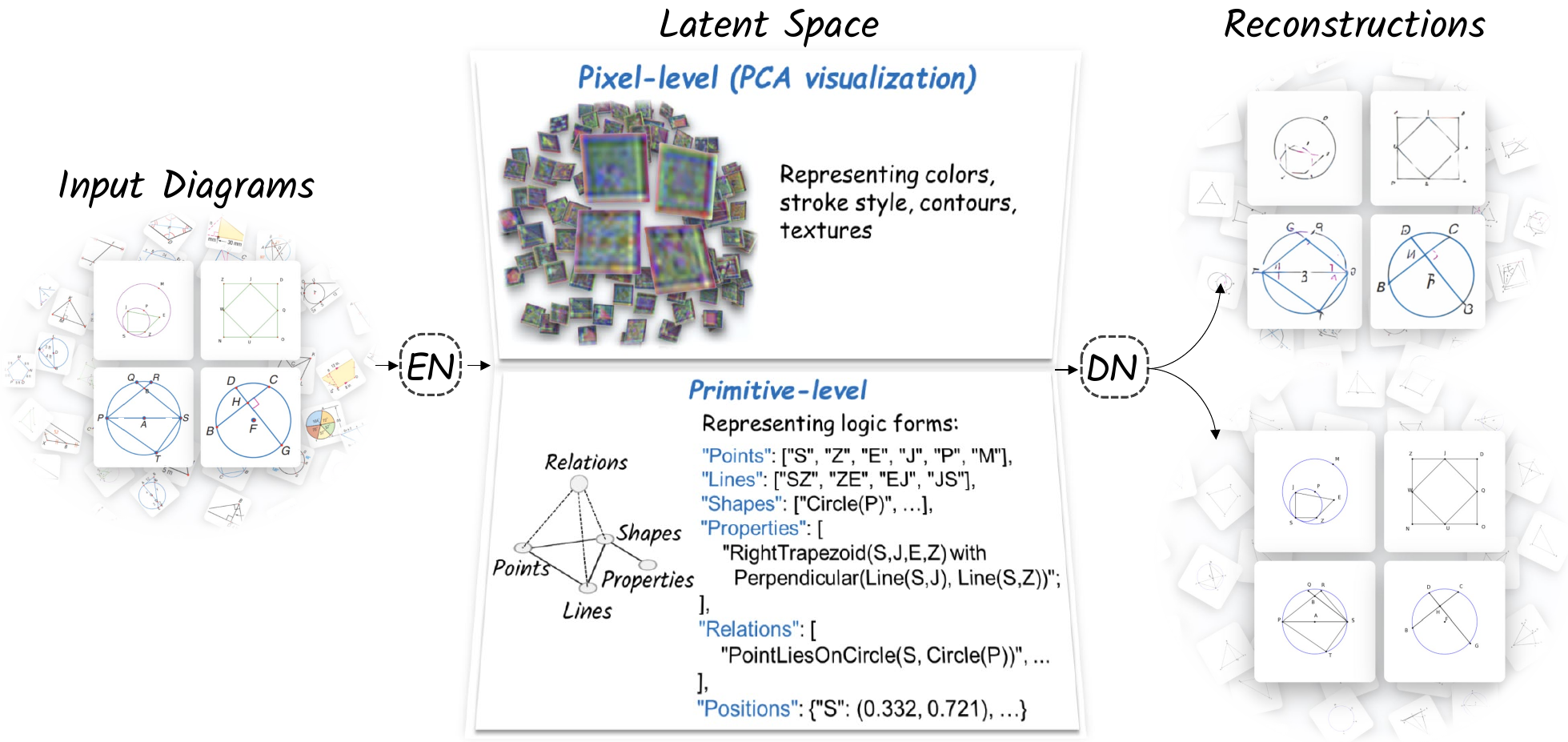}
    \vspace{-0.2cm}
    \caption{Comparison of latent spaces formed by semantic auto-encoder and our symbolic auto-encoder. Semantic feature vectors capture color and texture, which are uninformative for semantically sparse diagrams, leading to coarse-grained structural details (see Fig.~\ref{fig:recons_vis} for close-ups). Symbolic auto-encoder forms structured latent spaces, representing dependencies among primitives, with the decoder reconstructing diagrams based on visual-logic rules.}
    \vspace{-0.5cm}
    \label{fig:intro_fig}
\end{wrapfigure}
Unlike natural images that convey meaning through textures, colors, and scenes, diagrams are semantically sparse yet structurally rich blueprints. Mainstream vision encoders~\cite{radford2021learning,zhai2023sigmoid,oquab2023dinov2} treat diagram parsing as ordinary natural image processing, encoding diagrams into continuous high-dimensional vectors. Auto-encoder models~\cite{kingma2013auto,he2022masked} then reconstruct diagrams from these continuous feature embeddings, resulting in latent spaces that capture pixel-level semantics rather than symbolic structures. However, this paradigm is fundamentally limited for diagrams: when interpreting lines or arcs, pixel values carry no inherent semantic meaning, forcing models to approximate symbolic understanding from weak supervision signals. \textit{\textbf{What if the latent space itself were symbolic representations?}} We envision an auto-encoder whose encoder produces explicit symbolic graphs, \eg, mapping points to lines, lines to shapes, and their interrelationships, while the decoder functions as an executable engine that reconstructs diagrams directly from these primitives rather than from pixel values. Fig.~\ref{fig:intro_fig} shows such a proof-of-concept.


To achieve this goal, the key challenge lies in enabling the encoder to perceive not only individual primitives but also the hierarchical structures that are inherent and unique to diagrams. For instance, line detection must depend on correctly identified points; if lines are inaccurately parsed, subsequent shape analysis built upon them becomes meaningless. We introduce \ourrl{}(Symbolic Hierarchical Process Reward Modeling), which adapts Process Reward Models (PRMs) to provide fine-grained supervision for hierarchical multi-step parsing. Existing PRMs rely on either human annotation~\cite{uesato2022solving, lightman2024lets} (costly and unscalable) or LLM-as-a-judge~\cite{zheng2023judging} (prone to hallucination due to unreliable reward signals). \ourrl{} introduces three key innovations:  (1) \textbf{cost-free training data}, a synthetic logic-form engine that constructs structured parsing paths (\textsc{Point} $\Rightarrow$ \textsc{Line} $\Rightarrow$ \textsc{Shape} $\Rightarrow$ \textsc{Shape Properties} $\Rightarrow$ \textsc{Geometric Relations}) paired with rendered
diagrams, eliminating manual annotation; (2) \textbf{rule-based rewards} computed via verifiable metrics such as F1 scores and L2 distances, removing hallucinated supervision; (3) \textbf{hierarchical dependency modeling} that enforces step-level constraints, \eg, point-on-line, line-on-shape, and shape-on-relation.

After hierarchical reward training, we instantiate a symbolic auto-encoder with an executable rendering engine that reconstructs diagrams from structured primitives. This architecture enables self-supervised learning through perceptual loss between input diagrams and their reconstructions. However, diagram reconstruction fidelity offers sparse supervision, leading to reinforcement learning collapse—reward scores plateau early and KL divergences remains near zero (Fig.~\ref{fig:method_fig}). 
To address this, we introduce two stabilization strategies: (1) \textbf{hard negative contrastive learning} via Gaussian noise injection, increasing reward variance; and (2) \textbf{power normalization} of rewards, defined as $r' = r^\alpha$ with $\alpha > 1$, which amplifies reward differences while preserving their relative ranking.

We evaluate our model across three downstream tasks: diagram reconstruction, diagram understanding, and reasoning. Ablation studies reveal cross-domain generalization: it transfers primitives from planar geometry to solid-geometry attribution (\eg, color or material assignment) and aligns closely with natural language, translating logic-form rules into linguistic Chains-of-Thoughts, and extends to reconstructing electrical and chemical diagrams (\S~\ref{supp:recons} in Supp. Mat.). Beyond the conventional setup of integrating our vision model (visual tokens) with MLLM reasoners, we explore a neuro-symbolic system where symbolic primitives are passed directly to an LLM for multimodal reasoning. This pipeline enables visual rewards to adaptively complement textual answer rewards (typically final answer and format). We advocate future research toward neuro-symbolic architectures that can leverage the reasoning capabilities of deep learning while maintaining the logical rules and explainability of symbolic systems.

\section{Related Work}
\label{sec:related}


\noindent\textbf{Computer Vision Learners.}
Recent vision encoders~\cite{radford2021learning,oquab2023dinov2,li2022blip,zhai2023sigmoid} have demonstrated strong performance in natural image understanding, largely driven by vision–language contrastive learning. However, these models encode images into continuous embedding spaces optimized for semantic similarity rather than symbolic structure. When applied to diagrams, they struggle to capture geometric primitives and their relationships~\cite{wang2024mathvista,zhang2024mathverse,zhang2025primitive,sun2025mathglance,chen2024visiomath}. For self-supervised visual learning, Variational Autoencoders (VAEs)~\cite{kingma2013auto} and Masked Autoencoders (MAE)~\cite{he2022masked} define the auto-encoder paradigm. Self-Guided MAE~\cite{chen2024selfguided} improves masking via learned patch clustering. However, such methods form pixel-level latent spaces. In contrast, our work extends the auto-encoder paradigm to symbolic latent spaces, where the bottleneck is not continuous vectors but logic forms composed of explicit geometric primitives.

\noindent\textbf{Diagram Parsers.} Recent work on diagram parsing can be divided into code-based and logic-based representations. A growing line of research maps charts into executable Python code: ChartMimic~\cite{liu2024chartmimic} established an evaluation testbed, VisCodex~\cite{Jiang2025VisCodex} and ChartCoder~\cite{yan2025chartcoder} introduced large-scale chart–code pairs, and Plot2Code~\cite{wu2024plot2code} and PlotGen~\cite{zhang2025plotgen} further improved code executability and iterative refinement. While effective for statistical charts requiring numerical mapping, code-based representations are ill-suited for geometric diagrams. Libraries such as Matplotlib support only low-level primitives (points, circles) without explicit shapes or hierarchical relations, and even simple figures demand verbose, syntax-heavy code (\eg, \texttt{plt.subplots()}, styling boilerplate).  Logic-based approaches offer more structured representations. PGDP~\cite{zhang2022pgdp} parses diagrams into predicates such as \texttt{PointLiesOnLine} and \texttt{Perpendicular}, but lacks indicators of shape properties (\eg, explicitly marking an equilateral triangle with three equal sides) and captures only limited relations between points and lines. AlphaGeometry~\cite{trinh2024alphageometry} is a milestone in automated theorem proving, using clauses to solve IMO-level geometry problems. However, its representation is tailored for symbolic mathematical deduction rather than vision-grounded perception—it cannot parse or reconstruct visual diagrams. Our engine produces visual logic forms that encode primitive entities and shape-level information within a hierarchical structure, establishing the first self-supervised symbolic vision encoder with interpretable, compositional, and verifiable properties.

\noindent\textbf{Process Reward Models.} Process Reward Models (PRMs) for language reasoning~\cite{lightman2023lets} supervise intermediate reasoning steps rather than only final outcomes, enabling fine-grained credit assignment. Let’s Verify Step by Step shows that process supervision significantly improves mathematical problem solving, and GroundedPRM~\cite{zhang2025groundedprm} extends this with MCTS-guided search and tool-based verification. These works highlight the value of step-wise supervision but focus exclusively on textual reasoning. VisualPRM~\cite{wang2025visualprm} introduces an 8B-parameter process reward model for multimodal reasoning, and ~\cite{zhang2024training} explores test-time scaling via process supervision. However, these PRMs still operate on textual verification and lack explicit modeling of constraints among multisteps. In contrast, we introduce the first Hierarchical Process Reward Model (HPRM) for symbolic vision, supervising learning across multiple abstraction levels and incorporating explicit geometric dependencies. 




\section{Method}
\label{sec:method}

\subsection{Symbolic Encoder}

\begin{figure*}[t]
    \centering
    \includegraphics[width=\textwidth]{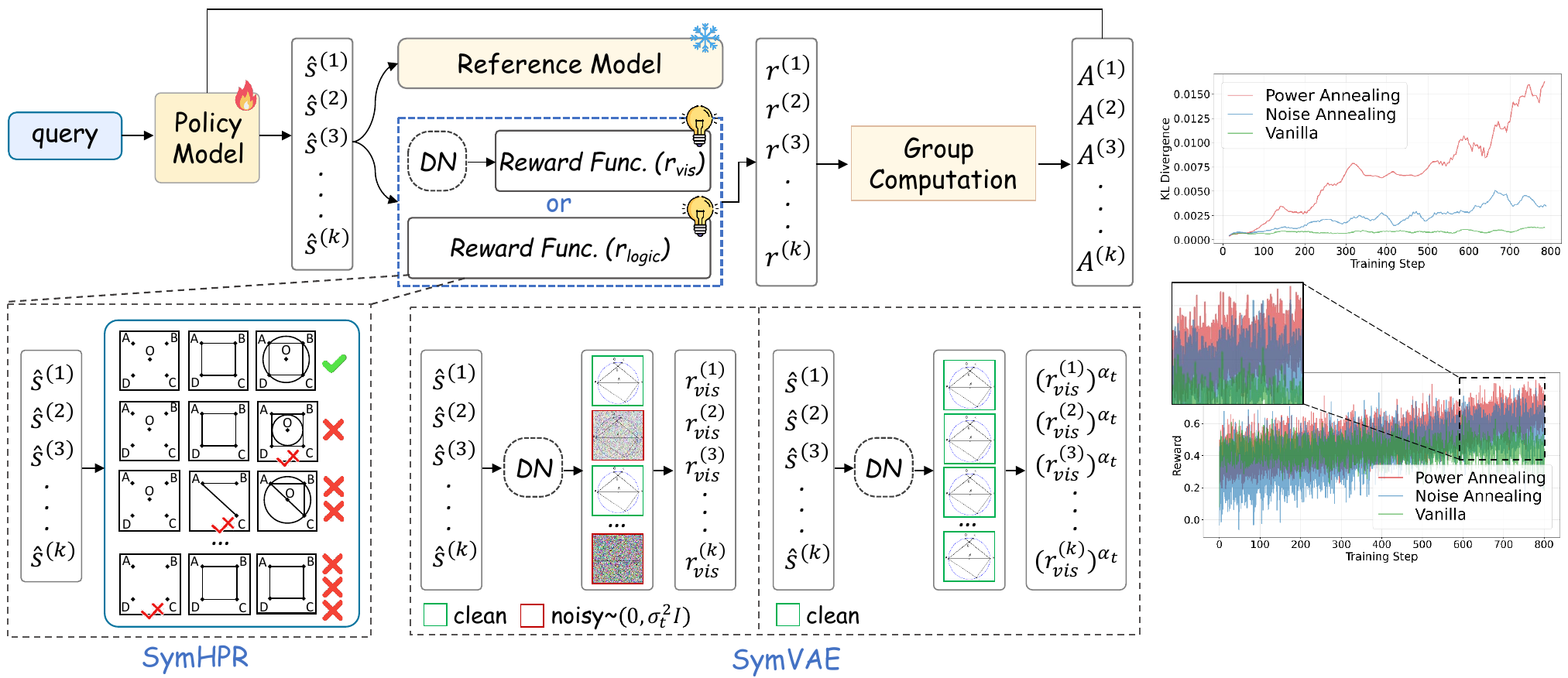}
    \vspace{-0.6cm}
    \caption{Overview of the symbolic auto-encoder training pipeline. \ourrl{} is optimized with hierarchical logic-form rewards to enforce compositional consistency across point–line–shape (the diagram decoding is shown here only for visualization). \ourvae{} employs Gaussian noise annealing (Eq.~\ref{eq:noise_anneal}) and power normalization (Eq.~\ref{eq:power_anneal}) to stabilize self-supervised training with visual rewards ($r_{\text{vis}}$). The effectiveness of stabilization techniques is demonstrated by the reward trajectories and KL divergence curves relative to the reference model.}
    \vspace{-0.5cm}
    \label{fig:method_fig}
\end{figure*}

\noindent\textbf{Problem Formulation.}
Let $\mathcal{X}$ denote the set of geometric diagram images and $\mathcal{S}$ the set of symbolic logic forms. We aim to learn an encoder $f_\theta: \mathcal{X} \to \mathcal{S}$ parameterized by a vision-language model $\theta$, which generates a structured symbolic representation $s \in \mathcal{S}$ from an input diagram $x \in \mathcal{X}$.

\noindent\textbf{Logic Forms.}
A logic form $s\!\! \in \! \! \mathcal{S}$ consists of compositional geometric primitives. Points are labeled with vertex letters (\eg, $a, b, c$) and associated with normalized spatial positions $(x, y) \in [0,1]^2$. Lines are defined by vertex pairs forming segments (\eg, $ab, bc$). Shapes are constructed from lines or circles, with special cases such as \texttt{RightTriangle} explicitly specifying which two lines are perpendicular. Geometric relations then describe interactions among primitives, including \texttt{IntersectAt}, \texttt{Incircle}, \texttt{Tangent}, and \texttt{AngleBisector}. This hierarchy captures how points define lines, lines form shapes, and those primitives collectively form higher-level relations. We denote ground-truth logic forms as $s$ and predictions as $\hat{s}$. Detailed examples are provided in the supplementary materials \S \ref{supp:logicform}.

\noindent \textbf{Cold-start Pre-training.} We first train the encoder using standard next-token prediction loss on paired data $\mathcal{D} = \{(x_i, s_i)\}_{i=1}^N$, enabling it to learn visual composition rules and the formal structure of the symbolic language:
\begin{equation}
\mathcal{L}_{\text{SFT}}(\theta) = -\mathbb{E}_{(x,s)\sim\mathcal{D}}\left[\log p_\theta(s|x)\right]
\label{eq:sft}
\end{equation}


\subsubsection{Hierarchical Process Rewards Modeling}

\noindent\textbf{Group Relative Policy Optimization.} After cold-start pre-training, we adopt reinforcement learning via GRPO~\cite{shao2024deepseekmath} to further optimize the encoder with fine-grained process rewards, enabling the model to explicitly capture the hierarchical relationships and constraints inherent to primitive elements. For each input diagram $x$, we sample $K$ logic-form candidates 
$\{\hat{s}^{(k)}\}_{k=1}^K$ from the current policy $p_{\theta}(s|x)$.  Each candidate is assigned a reward $r^{(k)}$, computed as the aggregate of task-specific perceptual reward functions to enhance policy learning.  The GRPO advantage for $k$-th sample is: $A^{(k)} = r^{(k)} - \frac{1}{K}\sum_{j=1}^K r^{(j)}$.

The policy gradient objective incorporates clipped importance sampling and KL divergence regularization:
\begin{equation}
\!\!\!\!\!\mathcal{L}_{\text{RL}}(\theta) = \mathbb{E}\Big[\min_{\theta}\big(r_t A^{(k)}, \text{clip}(r_t, 1-\epsilon, 1+\epsilon) A^{(k)}\big) - \beta \, \text{KL}[\pi_{\theta} \| \pi_{\text{ref}}]\Big]
\end{equation}

Here,  $r_t = \frac{p_\theta(\hat{s}|x)}{p_{\theta_{\text{old}}}(\hat{s}|x)}$ is the probability ratio, $A^{(k)}$ is the group-based advantage, $\epsilon$ controls the clipping range, and $\beta$ balances KL regularization. We use $K=8$ rollouts per prompt and $\beta=0.03$.  

{Task-specific reward functions} evaluate six geometric dimensions by comparing $\hat{s}$ with ground truth $s$, with reward scores normalized to $[0,1]$ via min–max scaling.

\noindent(1) Point reward is computed as the F1 score:
\begin{equation}
r_{\text{point}} = \frac{2 \cdot \text{Precision} \cdot \text{Recall}}{\text{Precision} + \text{Recall}} = \frac{2|\hat{s}_{\text{point}} \cap s_{\text{point}}|}{|\hat{s}_{\text{point}}| + |s_{\text{point}}|}
\end{equation}

\noindent (2) Line reward is evaluated using F1 score with order normalization, \eg, $ab \equiv ba$, as lines are not sensitive to vertex order: $r_{\text{line}} =
F_1(\hat{s}_{\text{line}}, s_{\text{line}})$.




\noindent (3) Shape reward requires matching both the shape type and its constituent vertices; for example, \texttt{Square(a,b,c,d)} matches only \texttt{Square(a,b,c,d)} (no
  permutations allowed): $r_{\text{shape}} = F_1(\hat{s}_{\text{shape}}, s_{\text{shape}})$.

\noindent (4) Shape indicator reward is conditioned on correctly predicted shapes, ensuring that indicators (\eg, right angles, parallel/perpendicular/equal sides) are credited only when the corresponding shape is correctly identified; formally, $r_{\text{indicator}} = r_{\text{shape}} \times F_1(\hat{s}_{\text{indicator}}, s_{\text{indicator}})$.

\noindent  (5) Geometric relation reward uses the F1 score computed on relation type-word only (\eg, extracting \texttt{PointLiesOnLine} from \texttt{PointLiesOnLine(n, Line(a,b))}), abstracting away the specific objects (already evaluated in shape-related rewards); formally, $r_{\text{relation}} = F_1(\hat{s}^{*}_{\text{relation}},s^{*}_{\text{relation}})$, where $*$ denotes filtering.

\noindent (6) Point-position reward measures the average Euclidean distance between predicted and ground-truth coordinates $(x,y)$ for correctly detected points, as follows:
\begin{align}
& d_{\text{avg}} = \frac{1}{|\hat{s}_{\text{point}} \cap s_{\text{point}}|} \sum_{i \in \hat{s}_{\text{point}} \cap s_{\text{point}}} \|\hat{p}_i - p_i\|_2, \nonumber \\
& r_{\text{position}} = \exp\left(-\frac{d_{\text{avg}}}{\tau}\right),
\end{align}
where $\tau = 0.05$ is the distance threshold for normalization.





\noindent\textbf{Hierarchical Reward Designs.} A naive approach aggregates the six reward components by simple averaging: $r_{\text{flat}} = \frac{1}{6}(r_{\text{point}} +
r_{\text{position}} + r_{\text{line}} + r_{\text{shape}} + r_{\text{indicator}} + r_{\text{relation}})$. However, this ignores the inherent compositional structure of
geometric primitives. We empirically find that under the flat mode, the policy model maintains performance similar to the cold-start model, indicating limited exploratory ability within the search space (Tab.~\ref{table:perception}).
We therefore explicitly model the hierarchical dependencies among reward components, enabling the policy to correctly account for the structural relationships among geometric primitives.

Let $\alpha_{c \rightarrow p} \in [0, 1]$ denote the dependency weight of component $c$ on its parent $p$. The adjusted reward for component $c$ is computed via blending:
\begin{equation}
r'_c = r_c \cdot \left[(1 - \alpha_{c \rightarrow p}) + \alpha_{c \rightarrow p} \cdot r'_p\right],
\end{equation}
where $\alpha = 0$ denotes full independence, $\alpha = 1$ denotes full dependency, and $0 < \alpha < 1$ indicates partial blending.
We set $\alpha = 0.9$ by default.
Dependencies are applied to the children of \textit{line}, \textit{shape}, and \textit{relation}, whose parents are \textit{point}, \textit{line}, and \textit{shape}, respectively.
We exclude the shape-indicator reward because it is already conditioned on correct shape prediction.
  \begin{equation}
  r'_c = \begin{cases}
  r_c & c \in \{\text{point, position, indicator}\} \\
  r_c \cdot [0.1 + 0.9 \cdot r'_p] & c \in \{\text{line, shape, relation}\}
  \end{cases}
  \label{eq:logic}
  \end{equation}
Finally, we compute the overall logic-form reward as the average of the adjusted components: $r_{\text{logic}} = \frac{1}{6}\sum_{c} r'_c$. This hierarchical structure ensures that geometric rewards are granted only when lower-level detections are reliable, preventing reward hacking through superficial matching.

\subsection{Rendering Engine as Decoder}
\label{sec:decoder}

To enable a self-supervised auto-encoding pipeline, we require a decoder that reconstructs diagrams from symbolic logic forms.
We define a deterministic rendering engine 
$R:\mathcal{S}\!\to\!\mathcal{X}$ 
that converts a logic form $s$ into a diagram image $\hat{x} = R(\hat{s})$.
The engine parses the structured logic form and renders the diagram using standard graphics libraries (\eg, \texttt{matplotlib}).
Implementation details are provided in the supplementary materials \S \ref{supp:decoder}.

\noindent\textbf{Differentiable Training via RL.} While the rendering function $R$ itself is non-differentiable, it enables gradient flow through reinforcement learning. The visual reward $r_{\text{vis}}(x, R(\hat{s}))$ provides a scalar supervision signal that guides policy updates via GRPO.  By reconstructing diagrams from symbolic geometric primitives and relations, this mechanism forces the encoder to produce logic forms that faithfully capture the original diagram's visual content, thereby enabling fully self-supervised training on unlabeled diagram images with a primitive-level latent space.



\subsubsection{Visual Reward Functions}

Following the auto-encoder paradigm~\cite{kingma2013auto}, we compute visual rewards by comparing the original diagram $x$ with the reconstructed image $\hat{x} = R(\hat{s})$. We employ three complementary reward components:

\noindent{(1) Reconstruction reward (MSE)} measures point-level fidelity via mean squared error: $ r_{\text{MSE}} = \text{MSE}(\hat{x}, x)$

\noindent{(2) Structural similarity (SSIM)} evaluates perceptual structural similarity:$ r_{\text{SSIM}} = \text{SSIM}(\hat{x}, x)$.

\noindent{(3) DINO similarity} measures cosine similarity between DINOv2~\cite{oquab2023dinov2} features of $\hat{x}$ and $x$:
$r_{\text{DINO}} = \cos\!\big(f_{\text{DINO}}(\hat{x}),\, f_{\text{DINO}}(x)\big)$.





\noindent The combined visual reward is a normalized weighted sum:
\begin{equation}
    r_{\text{vis}} = \sum_{k} \frac{w_k}{\sum_j w_j} r_k, \quad k \in \{\text{MSE, SSIM, DINO}\},
    \label{eq:vis}
\end{equation}
where $w_{\text{MSE}} = 0.6$, $w_{\text{SSIM}} = 0.3$, $w_{\text{DINO}} = 0.1$ by default.

\subsubsection{Stabilized GRPO for Symbolic Auto-Encoding}
\label{main:stabel}

We empirically find that directly applying standard GRPO with visual reward functions leads to training collapse due to small advantages (curves in Fig.~\ref{fig:method_fig}). This arises because diagrams contain large uniform backgrounds and simple foreground elements (lines/arcs), causing the visual reward functions to poorly differentiate reconstruction fidelity. As a result, reward variance becomes negligible, producing near-zero advantages and triggering policy collapse. Symptoms include: (1) reward scores plateau without improvement across training steps, and (2) KL divergence remains low, indicating the policy fails to explore the state space. We provide two solutions to address this critical challenge.

\noindent\textbf{Hard Negative Contrastive Learning.} To increase reward variance and strengthen the policy's learning, we apply Gaussian noise to reconstructed images during reward computation, creating hard negative samples. For each group of $K=8$ rollouts per prompt, 4 rollouts receive noisy reconstructions while the last 4 use clean reconstructions:
\begin{equation}
    \hat{x}_{\text{noisy}} = \text{clip}\left(\hat{x} + \mathcal{N}(0, \sigma^2 \mathbf{I}), 0, 255\right)
\end{equation}
Such noise injection creates a sharper reward distribution, with noisy rollouts receiving lower visual rewards $r_{\text{vis}}$ and clean samples achieving higher scores, thereby amplifying advantage magnitudes. This process resembles hard negative contrastive learning. However, using hard negatives throughout training would impair learning of useful geometric cues due to insufficient exploitation. We therefore employ progressive noise annealing to balance exploration and exploitation.
Specifically, we decay $\sigma$ exponentially during training to implement curriculum learning (exploration $\to$ exploitation). At training step $t$, the noise standard deviation is annealed as:
\begin{equation}
   \sigma_t = \sigma_{\min}^{(0)} + (\sigma_{\max}^{(0)} - \sigma_{\min}^{(0)}) \cdot \exp(- t / T),
\label{eq:noise_anneal}
\end{equation}
where $T = T_{\max}/2$ is half the total training steps. Noise is disabled when $\sigma_t < 0.01$, transitioning the policy from aggressive exploration (high noise $\sigma_{\max}=1.0$, early training) to stable exploitation (low noise $\sigma_{\min}=0.0$, late training).

\noindent\textbf{Power Normalization Annealing.} A principled way to sharpen the reward distribution is to apply power normalization. For any two rollouts with rewards $r_i > r_j$ and power $\alpha \geq 1$, the ratio is amplified under the transformation: $\frac{r_i^{\alpha}}{r_j^{\alpha}} > \frac{r_i}{r_j}$.
Exponentiating rewards increases the relative weight on higher-reward rollouts while decreasing the weight on lower-reward rollouts. We apply this principle by transforming group-normalized rewards via a power function:
  \begin{equation}
      \tilde{r}_i = \left( \frac{r_i - r^{\text{group}}_{\min}}{r^{\text{group}}_{\max} - r^{\text{group}}_{\min} + \epsilon} \right)^{\alpha_t},
      \label{eq:power_anneal}
  \end{equation}
where $r^{\text{group}}_{\min}$ and $r^{\text{group}}_{\max}$ denote the minimum and maximum rewards among the $K$ rollouts sampled for the same prompt.  Analogous to noise annealing in Eq.~\ref{eq:noise_anneal}, we decay $\alpha_t$ from $p_{\max}=3.0$ (aggressive early exploration) to $p_{\min}=1.0$ (stable late-stage training).
Larger powers exponentially amplify the differences between good and bad rollouts, yielding stronger and more informative learning signals for policy gradient updates.



Through ablation studies and monitoring of training curves (reward progression and KL divergence), we find that both techniques effectively
stabilize GRPO training. We adopt power normalization as the default configuration due to its superior final performance and training stability.
\section{Expriments}
\label{sec:exp}
\label{exp:main_results}

\begin{table}[!t]
\small
\centering
\caption{Performance comparison on geometric diagram reconstruction. We evaluate on two test sets: synthetic diagrams with high complexity, and  Geo170K real-world testset samples. We report Mean Squared Error (MSE, $\times 10^{-3}$) $\downarrow$, Learned Perceptual Image Patch Similarity (LPIPS) $\downarrow$,  Structural Similarity Index (SSIM) $\uparrow$, and DINO feature similarity $\uparrow$. $\ominus$ denotes vanilla GRPO, and $\oplus$ indicates the noise-annealing strategy.}
\vspace{-.5em}
\label{tab:geometric_reconstruction}
\renewcommand\arraystretch{1}
\setlength{\tabcolsep}{3mm}{{
\begin{tabular}{l|cccc|cccc}
\hline
\multirow{2}{*}{\textbf{Model}} & \multicolumn{4}{c|}{\textbf{Synthetic Diagrams}} & \multicolumn{4}{c}{\textbf{Geo170K Diagrams}} \\
\cline{2-9}
& \textbf{MSE}$\downarrow$ & \textbf{LPIPS}$\downarrow$ & \textbf{SSIM}$\uparrow$ & \textbf{DINO}$\uparrow$ & \textbf{MSE}$\downarrow$ & \textbf{LPIPS}$\downarrow$ & \textbf{SSIM}$\uparrow$ & \textbf{DINO}$\uparrow$  \\
\hline
\rowcolor{green!15}\multicolumn{9}{l}{\textit{Pixel-Level Auto-Encoder}}   \\ \hline
VAE~\cite{kingma2013auto}& 12.9& 0.29 & 0.62 & 0.81 & 37.9& 0.37 & 0.64 & 0.80 \\
VQ-GAN~\cite{tamingtransformershighresolutionimage} & 11.7 & 0.21 & 0.69 & 0.81 & 37.2 & 0.31 & 0.73 & 0.83\\
\hline
\rowcolor{green!15}\multicolumn{9}{l}{\textit{Open/Close-Source MLLMs}}\\\hline
GPT-4o & 34.3 & 0.15 & 0.82 & \textbf{0.96} & 39.9 & 0.22 & 0.76 & 0.92\\
GPT-o3 & 651.4 & 0.29 & 0.28 & 0.77 & 395.4 & 0.27 & 0.46 & 0.83\\
\hline
Qwen2.5-VL-3B & 167.8 & 0.30 & 0.62 & 0.83 & 109.2 & 0.31 & 0.64 & 0.83 \\
Qwen2.5-VL-7B & 703.0 & 0.32 & 0.22 & 0.74 & 667.3 & 0.36 & 0.22 & 0.72 \\ 
\hline
\rowcolor{green!15}\multicolumn{9}{l}{\textit{Python-Code}}   \\ \hline
PyhParser-3B & 8.98 & 0.18 & 0.71 & 0.86 & 38.9 & 0.32 & 0.71 & 0.86\\
\hline
\rowcolor{green!15}\multicolumn{9}{l}{\textit{Symbolic-Logics}}   \\ \hline
\ourparser-3B & 7.64 & 0.12 & 0.76 & 0.93& 36.8 & 0.28 & 0.76 & 0.93 \\
\ourrl-3B & 7.02 & 0.08 & 0.83 &\textbf{0.96} & 27.7 & 0.26 & 0.77 & 0.95 \\
\ourvae-3B$\ominus$ &  6.99 & 0.10 & 0.81 &0.91 & 28.3 & 0.26 & 0.75 & 0.94\\
\ourvae-3B$\oplus$ & 6.20 &0.06 & 0.87 & 0.92 & 22.4 & 0.21& 0.78 & 0.95 \\
\ourvae-3B & 6.13& \textbf{0.01} & 0.89 & 0.95 & 21.8 & 0.20& 0.79 & 0.95 \\
\hline
\bf \ourvae-7B & \textbf{6.01}& 0.05 & \textbf{0.94} & \textbf{0.96} & \textbf{16.8} & \textbf{0.17} & \textbf{0.83} & \textbf{0.96} \\
\hline
\end{tabular}}}
\label{table:recons}
\vspace{-0.4cm}
\end{table}

\subsection{Implementation Details}
\noindent\textbf{Training Pipeline.} We adopt a three-stage training pipeline for symbolic auto-encoder (\ourvae). We first perform cold-start training, enabling the base model to predict structured logic forms (\ourparser). We then apply Hierarchical Process Reward Modeling (\ourrl) to enforce compositional dependencies among geometric primitives. For datasets, we use 100K / 9K synthetic diagram–logic-form pairs (synthetic generation described in \S\ref{supp:datasyn}) and 16K diagram-only samples (7K synthetic diagrams+5K Geo170K~\cite{gao2023gllava}+4K PGDP~\cite{zhang2022pgdp}). The training strategy is summarized in Tab.~\ref{tab:training_pipeline}. 

\begin{wraptable}{r}{9.5cm}
\begin{minipage}{1.0\linewidth}
\small
\centering
\vspace{-0.6cm}
\caption{Training settings for our variants.}
\label{tab:training_pipeline}
\vspace{-0.3cm}\setlength{\tabcolsep}{0.1mm}
{
\begin{tabular}{lccc}
\toprule
\textbf{Variants} & \textbf{Base Model} & \textbf{Dataset/Size} & \textbf{Training Type} \\
\midrule
\ourparser & Qwen2.5-VL & Diagrams+logic forms/100K & SFT (Eq.~\ref{eq:sft}) \\
\ourrl & \ourparser & Diagrams+logic forms/9K & RL (Eq.~\ref{eq:logic}) \\
\ourvae & \ourrl &  Diagrams/16K & RL (Eq.~\ref{eq:vis}) \\
\bottomrule
\end{tabular}
}
\vspace{-0.3cm}
\label{tab:training_pipeline}
\end{minipage}
\end{wraptable}

During downstream fine-tuning, we adapt the model to specific tasks using LoRA~\cite{hu2021lora} (rank = 64). For diagram understanding, we fine-tune it on 10K perception samples (one-tenth of the MathGlance training dataset, GeoPeP~\cite{sun2025mathglance}). For geometric reasoning, following prior work~\cite{zhang2025primitive,zhang2024mavis} that swaps visual encoders into off-the-shelf MLLM reasoners, we replace the visual encoder in Qwen2.5-VL-7B with ours and train with Chain-of-Thoughts supervision on Geo170K~\cite{gao2023gllava} and MathV360K~\cite{shi2024mathllava}.
To evaluate how the symbolic vision encoder benefits chart parsing, we further fine-tune our model on 1/5 of the VisCodex dataset~\cite{Jiang2025VisCodex} (100K chart–code pairs). Full training details, including optimizer settings, GRPO hyperparameters, and infrastructure configurations, are provided in \S\ref{supp:traindetail}.

\noindent\textbf{Evaluation Benchmarks.}
\textit{Reconstruction:} We evaluate on 500 synthetic and 500 Geo170K test diagrams using MSE, SSIM, LPIPS~\cite{zhang2018unreasonable}, and DINO~\cite{oquab2023dinov2} similarity metrics. 
For chart reconstruction, we use the ChartMimic Direct Mimic benchmark~\cite{liu2024chartmimic} with both low- and high-level scores.  \textit{Diagram understanding:} We report top-1 accuracy on the MathGlance benchmark (planar/solid geometry and graphs)~\cite{sun2025mathglance}.  \textit{Reasoning:} We evaluate answer accuracy and step-level correctness on MathVerse~\cite{zhang2024mathverse} and GeoQA~\cite{chen2021geoqa}. Additional details are in \S\ref{supp:traindetail}.

\subsection{Main Results}


\begin{table*}[t]
\footnotesize
\centering
\vspace{-0.5em}
\begin{minipage}[t]{0.48\linewidth}
\centering
\renewcommand\arraystretch{1.2}
\setlength{\tabcolsep}{2mm}
\caption{Top-1 accuracy (\%) on the MathVerse \texttt{test-mini} set. Models are evaluated across two visual dependency levels: \textit{Vision-Intensive} (requires visual understanding) and \textit{Vision-Only} (purely visual tasks). Close-up results are provided in \S\ref{supp:expresults}.}
\label{table:mathverse}
\begin{tabular}{l|c|c|c}
\hline
\textbf{Model} & \textbf{All} & \textbf{Vision Int.} & \textbf{Vision Only} \\
\hline
\rowcolor{green!15}\multicolumn{4}{l}{\textit{Baselines}}   \\ \hline
Random Chance & 12.4 & 12.4 & 12.4 \\
Human & 67.7 & 61.4 & 66.7 \\
\hline
\rowcolor{green!15}\multicolumn{4}{l}{\textit{Open-Source MLLMs}}   \\ \hline
SPHINX-MoE & 15.0 & 14.8 & 9.1 \\
G-LLaVA-7B & 16.6 & 17.2 & 9.4 \\
LLaVA-1.5-7B & 7.6 & 7.4 & 6.9 \\
ShareGPT4V & 13.1 & 15.5 & 3.7 \\
Math-LLaVA-7B & 19.0 & 20.2 & 16.4 \\
MAVIS-7B & 28.4 & 24.7 & 18.3 \\
Math-LLaVA-13B & 24.1 & 17.6 & 16.4 \\
MutilMath-7B & 26.9 & 25.9 & 15.0 \\
Primitive-7B & 24.3 & 25.6 & 17.5 \\
LLaVA-NeXT-110B & 28.3 & 22.1 & 20.7 \\
Qwen2.5-VL-7B & 49.2 & 33.2 & 21.1 \\
\hline
\rowcolor{green!15}\multicolumn{4}{l}{\textit{Our Models}}   \\ \hline
\bf \ourvae+CoTs-7B & \textbf{51.8} & \textbf{35.2} & \textbf{24.9} \\
\hline
\end{tabular}
\end{minipage}
\hfill
\begin{minipage}[t]{0.48\linewidth}
\centering
\renewcommand\arraystretch{1.34}
\setlength{\tabcolsep}{4mm}
\caption{Performance comparison on GeoQA. We report numerical answer accuracy (\%) for our \textbf{7B} models. The neuro-symbolic pipeline: \ourvae{} converts diagrams into executable logic forms, which are prepended to the question and processed by LLM (Qwen2.5-Math-7B).}
\label{table:geoqa}
\begin{tabular}{l|c}
\hline
\textbf{Model} & \textbf{Accuracy (\%)} \\
\hline
\rowcolor{green!15}\multicolumn{2}{l}{\textit{Baselines}}   \\ \hline
Random Chance & 25.0 \\
Frequent Guesses & 32.1 \\
\hline
\rowcolor{green!15}\multicolumn{2}{l}{\textit{Open-Source MLLMs}}   \\ \hline
G-LLaVA-7B & 64.2 \\
MAVIS-7B & 66.7 \\
Primitive-7B & 67.0 \\
Math-LLaVA-13B & 60.7 \\
MutilMath-7B & 74.1 \\
Qwen2.5-VL-7B & 76.4 \\
\hline
\rowcolor{green!15}\multicolumn{2}{l}{\textit{Our Models}}   \\ \hline
Logic forms $\xrightarrow{\text{w/o } r_{\text{vis}}} \text{LLM}$ & 63.4 \\
\bf Logic forms $\xrightarrow{\text{w } r_{\text{vis}}} \text{LLM}$ & 67.9 \\
\hline
\ourvae+CoTs (w proj.) & 79.2 \\
\bf \ourvae+CoTs (w/o proj.) & \textbf{79.4} \\
\hline
\end{tabular}
\end{minipage}
\vspace{-0.3cm}
\end{table*}

We evaluate our symbolic auto-encoder (\ourvae) across three tasks: diagram reconstruction, diagram understanding, and mathematical reasoning. Additional reconstruction examples, \eg, electric and chemical diagrams, are in \S~\ref{supp:recons}.

\noindent\textbf{Geometric Diagram Reconstruction.}
We evaluate on two test sets: (1) synthetic diagrams with complex structures and (2) Geo170K diagrams~\cite{gao2023gllava} (unseen styles excluded from training \ourparser{} and \ourrl). \ourvae-7B achieves the best performance across all metrics, demonstrating superior reconstruction fidelity in both complex and real-world scenarios. Further analysis is in Sec.~\ref{exp:abla}.

\begin{wraptable}{r}{9.5cm}
\begin{minipage}{1.0\linewidth}
\small
\centering
\vspace{-1.5em}
\caption{Performance comparison on ChartMimic Direct Mimic task for chart reconstruction.} 
\label{tab:chart_reconstruction}
\renewcommand\arraystretch{1}
\setlength{\tabcolsep}{1.5mm}{
\begin{tabular}{l|ccc|c}
\hline
\multirow{2}{*}{\textbf{Model}} & \multicolumn{3}{c|}{\textbf{Low-Level}} & \textbf{High-Level} \\
\cline{2-4}
& \textbf{Layout} & \textbf{Type} & \textbf{Avg.} & \textbf{GPT-4V} \\
\hline
\rowcolor{green!15}\multicolumn{5}{l}{\textit{Closed-Source MLLMs}}   \\ \hline
GPT-4o & 89.8 & \textbf{77.3} & 79.0 & \textbf{83.5} \\
Claude-3-opus & 83.1 &49.9 & 60.5 & 60.1 \\
GeminiProVision & 74.5& 49.2 & 54.8 & 62.2\\
\hline
\rowcolor{green!15}\multicolumn{5}{l}{\textit{Open-Source MLLMs}}   \\ \hline
InternVL2-8B & 51.1 & 28.6 & 34.4 & 38.9 \\
Qwen2-VL-7B~\cite{bai2023qwenvl} & 51.0& 31.0 & 32.9 & 35.0 \\
Qwen2.5-VL-7B~\cite{bai2023qwenvl} & 52.0& 30.3 & 33.2 & 34.2 \\
LLaVA-Next-Mistral-7B ~\cite{liu2024llavanext}& 31.1& 19.8 & 20.7 & 21.3\\
VisCodex-8B~\cite{Jiang2025VisCodex} & -- & --& 74.8 & 74.1 \\
\hline
\rowcolor{green!15}\multicolumn{5}{l}{\textit{Our Models}}   \\ \hline
\bf \ourvae+chart-7B & \textbf{96.3} & 76.8 & \textbf{79.6} & 77.3 \\
\hline
\end{tabular}}
\label{table:chart}
\vspace{-0.7cm}
\end{minipage}
\end{wraptable}

\noindent\textbf{Chart Reconstruction.}
We assess chart parsing using the ChartMimic Direct Mimic benchmark (Tab.~\ref{table:chart}). \ourvae+Chart-7B surpasses GPT-4o by +0.6\% on low-level metrics and notably improves layout accuracy by +6.5\%. Despite being trained on five times fewer chart samples, our model outperforms the specialized chart parser VisCodex-8B (74.8\%) and significantly exceeds the base Qwen2.5-VL-7B (33.2\%), highlighting the strong generalization capability of symbolic logic-form representations.

\noindent\textbf{Diagram Understanding.}
As shown in Tab.~\ref{table:perception}, \ourvae-7B achieves 72.6\% average accuracy across planar geometry, solid geometry, and graph analysis—an absolute improvement of 13.4\% over Qwen2.5-VL-7B (59.2\%) and outperforming all open-source MLLMs, \eg, Math-LLaVA-13B and Primitive-7B. The largest gains appear in relation identification (100.0\% \vs 52.0\% for Qwen2.5-VL-7B), where compositional reasoning over geometric primitives is essential. Notably, \ourvae-7B surpasses GPT-4o-mini-high by +2.9\% on solid geometry and +12.1\% on graph analysis.

\begin{table*}[!t]\footnotesize
 \caption{Performance comparison across plane geometry, solid geometry, and graphs in different tasks: shape classification, object counting, object grounding, and relationship identification (\textit{cls}, \textit{cnt}, \textit{grd}, and \textit{rlat}). \textit{all} indicates the overall accuracy, calculated as the ratio of correctly answered questions to the total number of questions, while \textbf{Avg.} denotes the average \textit{all} score across all subjects.}
 \vspace{-.5em}
 \label{tab:main_results}
 \centering
 \renewcommand\arraystretch{1.2}
 \setlength{\tabcolsep}{1.6mm}{
 \resizebox{1.0\linewidth}{!}{
\begin{tabular}{l|c|c|ccccc|ccccc|ccccc}
\hline
\multirow{2}{*}{\textbf{Model}} & \multirow{2}{*}{\textbf{Size}} & \multirow{2}{*}{\textbf{Avg.}} & \multicolumn{5}{c|}{\textbf{Plane  Geometry}} & \multicolumn{5}{c|}{\textbf{Solid  Geometry}} & \multicolumn{5}{c}{\textbf{Graphs}} \\ \cline{4-18} 
 &&& \textit{all} & \textit{cls} & \textit{cnt} & \textit{grd} & \textit{rlat} & \textit{all} & \textit{cls} & \textit{cnt} & \textit{grd} & \textit{rlat} & \textit{all} & \textit{cls} & \textit{cnt} & \textit{grd} & \textit{rlat} \\ \hline
\rowcolor{green!15}\multicolumn{18}{l}{\textit{Close-Source MLLMs}}   \\ \hline
GPT-4o & -& 53.3 & 42.8 & 58.4 & 53.2 & 1.1 & 62.5 & 60.7 & 72.1 & 84.5 & 1.6 & \textbf{66.3} & 56.4 & 92.8 & 72.2 & 1.6 & 57.6 \\ 
GPT-o1 & - & 36.5 & 15.8 & 33.2 & 11.6 & 0.0 & 14.0 & 41.4 & 75.6 & 52.6 & 0.0 & 23.8 & 52.3 & 82.6 & 81.5 & 0.0 & 39.4 \\ 
GPT-o4-mini-high & - & 48.0 & 19.1 & 29.3 & 24.4 & 0.4 & 21.5 & 64.7 & 94.2 & 79.4 & 0.0 & \textbf{66.3} & 60.1 & \textbf{95.7} & 77.8 & 0.0 & 69.7 \\  \hline
\rowcolor{green!15}\multicolumn{18}{l}{\textit{Open-Source MLLMs}}   \\ \hline 
LLaVA-v1.5~\cite{liu2023visual} & 7B & 33.3 & 29.2 & 29.0 & 39.6 & 14.2 & 37.5 & 31.6 & 43.0 & 42.3 & 0.0 & 31.3 & 39.0 & 76.8 & 35.2 & 0.0 & 39.4 \\ 
LLaVA-v1.5~\cite{liu2023visual} &13B & 35.4 & 32.8 & 29.3 & 40.4 & 23.5 & 42.0 & 35.9 & 60.5 & 38.1 & 0.0 & 35.0 & 37.6 & 63.8 & 42.6 & 0.0 & 45.5 \\  
Qwen2-VL~\cite{bai2023qwenvl} & 7B& 51.4 & 37.9 & 47.6 & 41.2 & 12.8 & 53.0 & 64.1 & 93.0 & 78.4 & \textbf{14.3} & 55.0 & 52.3 & 84.1 & 88.9 & 3.2 & 18.2 \\ 
Qwen2.5-VL~\cite{bai2023qwenvl}& 7B& 59.2 & 44.0 & 56.2 & 51.3 & 18.5 & 52.0 & 63.1 & \textbf{98.8} & 88.7 & 0.0 & 65.0 & 65.7 & 89.9 & \textbf{100.0} & 3.2 & 78.8 \\ 
DeepSeek-VL2-Tiny~\cite{wu2024deepseek}& 3B& 32.6 & 29.5 & 45.2 & 34.4 & 4.6 & 32.0 & 39.0 & 76.7 & 32.0 & 0.0 & 37.5 & 29.4 & 39.1 & 57.4 & 0.0 & 18.2 \\ 
 Math-LLaVA~\cite{shi2024mathllava}& 13B& 40.0 & 27.9 & 34.4 & 32.4 & 0.0 & 50.5 & 44.8 & 81.4 & 55.7 & 0.0 & 27.5 & 47.3 & 78.3 & 59.3 & 0.0 & 51.5 \\ 
G-LLaVA~\cite{gao2023gllava} &7B& 30.3& 25.6 & 27.8 & 41.2 & 0.4 & 38.0 & 32.3 & 45.4 & 38.1 & 4.8 & 32.5 & 33.9 & 58.0 & 37.0 & 0.0 & 42.4 \\ 
MultiMath~\cite{peng2024multimath}&7B & 42.1 & 31.2 & 44.0 & 30.4 & 1.1 & 53.0 & 46.7 & 81.4 & 53.6 & 4.7 & 33.8 & 48.6 & 79.7 & 57.4 & 3.2 & 33.8 \\ 
Primitive~\cite{zhang2025primitive} & 7B& 46.6 & 35.4 & 52.4 & 36.0 & 3.6 & 51.0 & 49.4 & 77.9 & 62.9 & 1.5 & 41.3 & 55.1 & 81.2 & 75.9 & 1.6 & 69.7 \\ \hline
\rowcolor{green!15}\multicolumn{18}{l}{\textit{Our Models}}   \\ \hline
PyhParser+GeoPeP & 3B& 11.4 & 2.77 & 3.13 & 4.52 & 0.0 & 3.41 & 31.5 & 93.0& 8.21& 0.0 & 24.6 &  0.0 & 0.0 & 0.0& 0.0 & 0.0 \\ 
\ourparser+GeoPeP & 3B& 65.4 & 69.9 & 84.2 & {83.3} & 21.4 & 90.5 &  60.8& 95.7 & 93.1& 0.0 & 54.3 &  65.5 & 88.3 & 97.1& 0.2 & 76.4 \\ 
\ourrl (flat.)+GeoPeP & 3B& 65.7 & 69.7 & 84.2 & {83.4} & 21.9 & 89.3 & 60.8 & 95.8 & 93.7& 0.0 & 53.7 &  66.7  &88.0 & 97.0& 0.0 & 81.8 \\ 
\ourrl+GeoPeP & 3B&  69.3 & 76.1 & 85.1 & 84.3 & 37.2 & 97.7 & 64.3 & 96.1 & 96.3& 1.3 & 63.3 &   67.5  & 90.1 & 97.2& 1.2 & 81.3 \\\hline
\bf \ourvae+GeoPeP & 7B&  \textbf{72.6} & \textbf{77.9} & \textbf{87.1} & \textbf{85.6} & \textbf{38.9} & \textbf{100.0} & \textbf{67.6} & 98.6 & \textbf{99.3}& 11.3 & 61.3 &  \textbf{72.2}  & 92.1 & \textbf{100.0} & \textbf{8.7} & \textbf{87.9} \\ 
\hline
\end{tabular}}}
\label{table:perception}
\vspace{-0.3cm}
\end{table*}

\begin{wrapfigure}{r}{10cm}
\begin{minipage}{1.0\linewidth}
    \centering
    \vspace{-0.6cm}
    \begin{subfigure}[t]{0.48\linewidth}
        \centering
        \includegraphics[width=\linewidth]{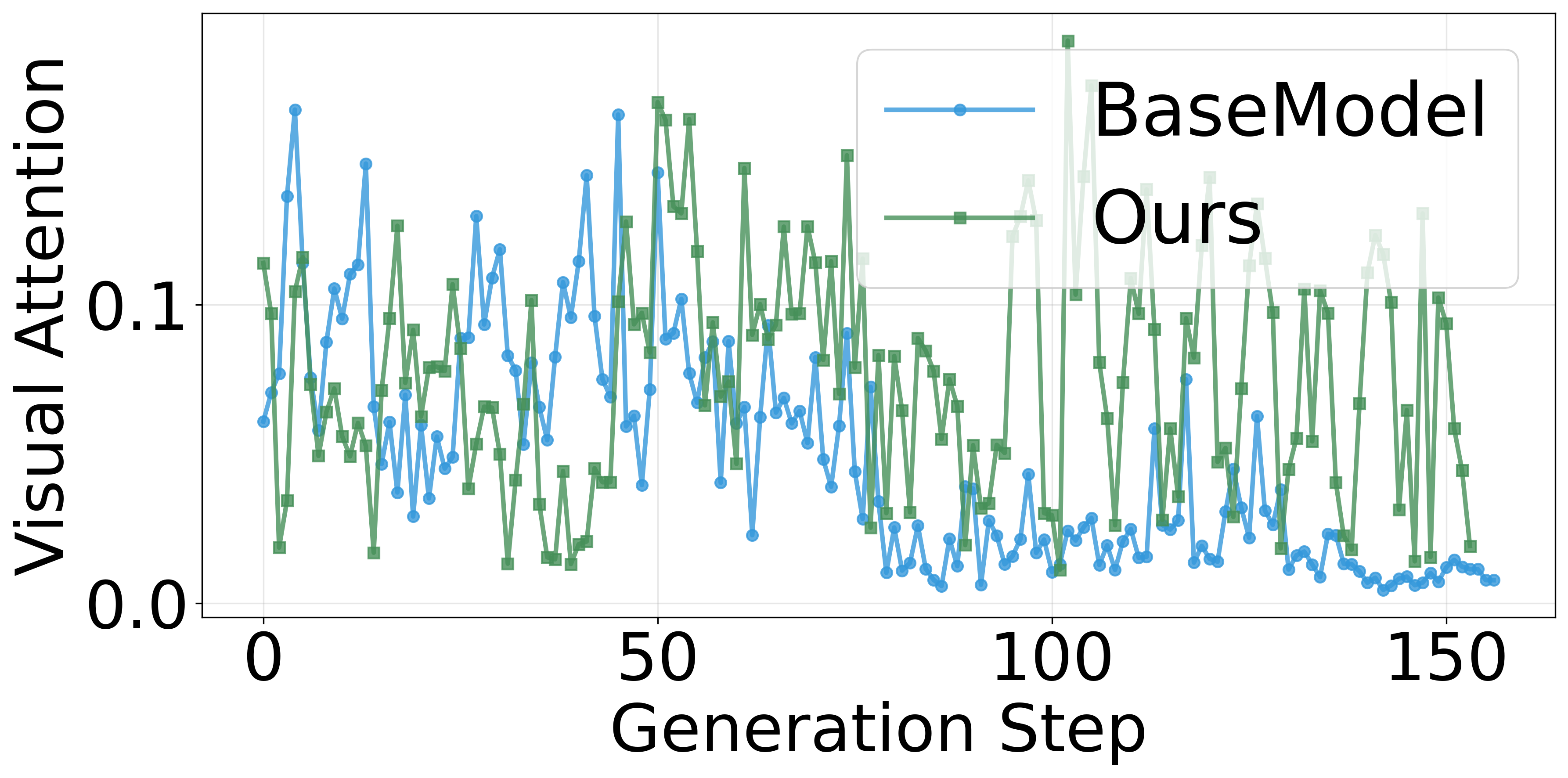}
        \label{fig:per_attention}
    \end{subfigure}
    \hfill
    \begin{subfigure}[t]{0.48\linewidth}
        \centering
        \includegraphics[width=\linewidth]{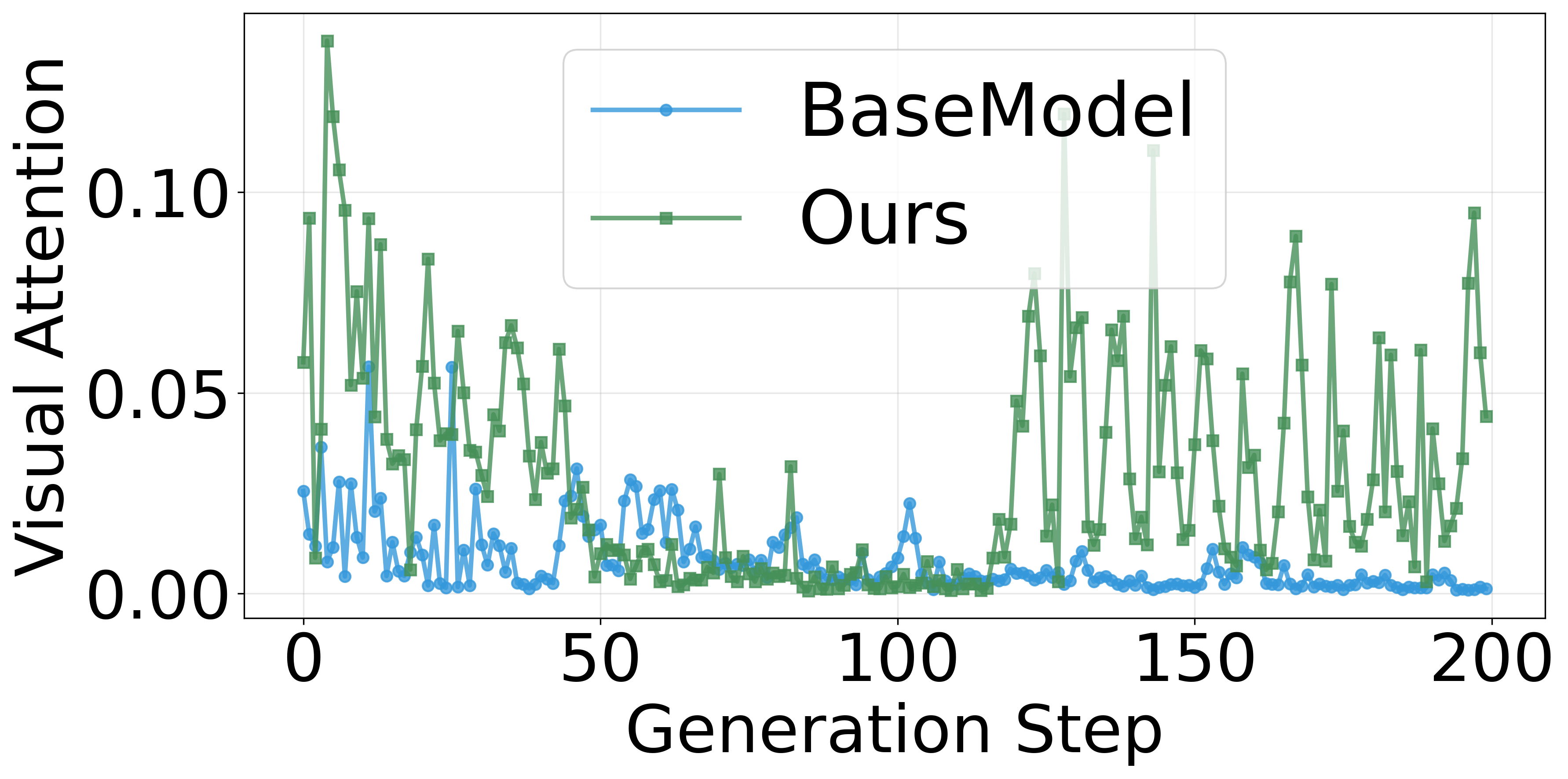}
        \label{fig:reason_attention}
    \end{subfigure}
    \vspace{-0.6cm}
    \caption{Cross-modal attention visualization during geometric description (left) and reasoning (right).}
    \label{fig:attention_comparison}
    \vspace{-0.4cm}
\end{minipage}
\end{wrapfigure}
\noindent\textbf{Mathematical Reasoning.}
For geometric problem solving, we evaluate on MathVerse and GeoQA benchmarks. On MathVerse (Tab.~\ref{table:mathverse}), our 7B \ourvae+CoT achieves 51.8\% overall accuracy, outperforming Qwen2.5-VL-7B (49.2\%), particularly in vision-dominant settings, indicating improved visual grounding through symbolic representations. To analyze the improvement, we visualize cross-modal attention between reasoning verbal tokens and visual tokens. The base model (Qwen2.5-VL-7B) shows weak visual grounding, whereas our model maintains consistent visual focus throughout reasoning. Even when prompted with diagram descriptions, the base model still fails to attend to visual content (Fig.~\ref{fig:attention_comparison}). On GeoQA, \ourvae+CoT reaches 79.4\% accuracy, surpassing Qwen2.5-VL-7B (76.4\%) and MultiMath-7B (74.1\%). Notably, our model without projector alignment (79.4\%) matches the performance of projector-based variants (79.2\%), suggesting that symbolic logic forms are inherently compatible with the language reasoning space, obviating the need for explicit visual–linguistic alignment (Tab.~\ref{table:geoqa}).

\begin{figure}[t]
\centering

\resizebox{0.6\linewidth}{!}{%
\begin{minipage}{\linewidth}
\centering

\begin{subfigure}[b]{0.24\linewidth}
    \centering
    \includegraphics[width=\linewidth]{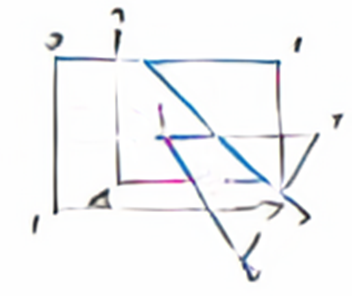}
\end{subfigure}
\hfill
\begin{subfigure}[b]{0.24\linewidth}
    \centering
    \includegraphics[width=\linewidth]{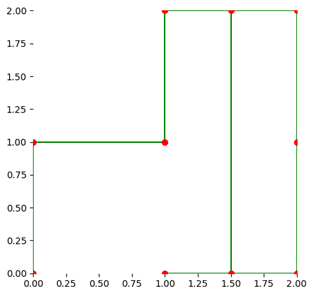}
\end{subfigure}
\hfill
\begin{subfigure}[b]{0.24\linewidth}
    \centering
    \includegraphics[width=\linewidth]{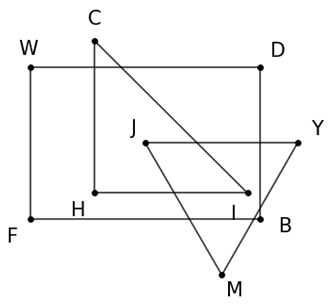}
\end{subfigure}
\hfill
\begin{subfigure}[b]{0.24\linewidth}
    \centering
    \includegraphics[width=\linewidth]{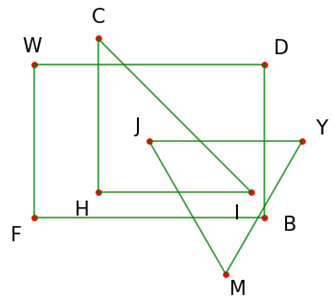}
\end{subfigure}

\begin{subfigure}[b]{0.24\linewidth}
    \centering
    \includegraphics[width=\linewidth]{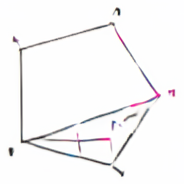}
\end{subfigure}
\hfill
\begin{subfigure}[b]{0.24\linewidth}
    \centering
    \includegraphics[width=\linewidth]{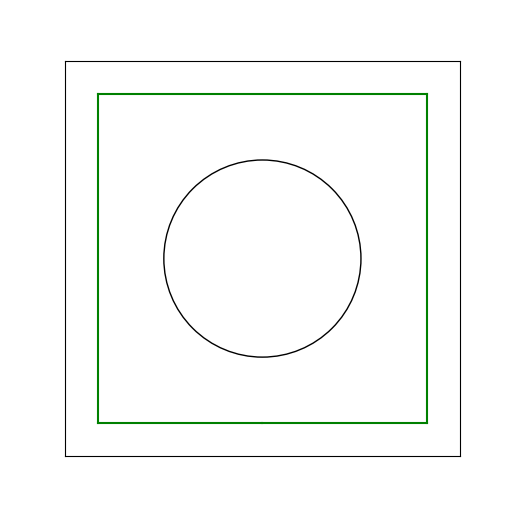}
\end{subfigure}
\hfill
\begin{subfigure}[b]{0.24\linewidth}
    \centering
    \includegraphics[width=\linewidth]{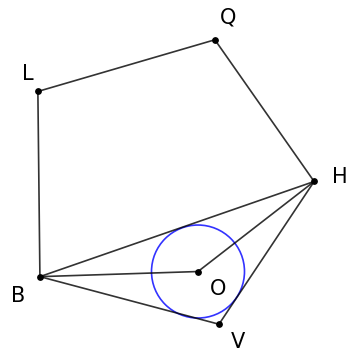}
\end{subfigure}
\hfill
\begin{subfigure}[b]{0.24\linewidth}
    \centering
    \includegraphics[width=\linewidth]{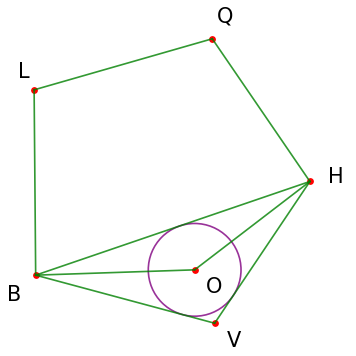}
\end{subfigure}

\begin{subfigure}[b]{0.24\linewidth}
    \centering
    \includegraphics[width=\linewidth]{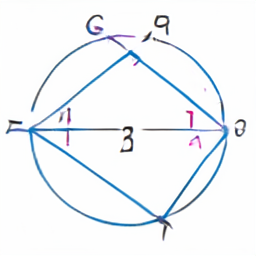}
\end{subfigure}
\hfill
\begin{subfigure}[b]{0.24\linewidth}
    \centering
    \includegraphics[width=\linewidth]{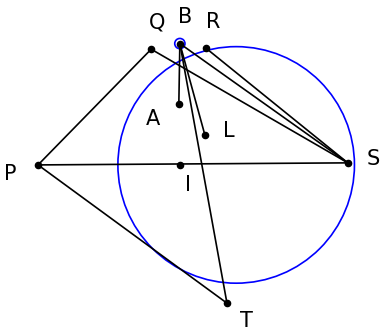}
\end{subfigure}
\hfill
\begin{subfigure}[b]{0.24\linewidth}
    \centering
    \includegraphics[width=\linewidth]{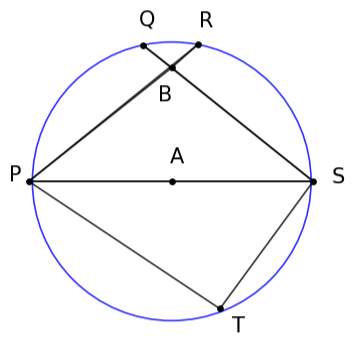}
\end{subfigure}
\hfill
\begin{subfigure}[b]{0.24\linewidth}
    \centering
    \includegraphics[width=\linewidth]{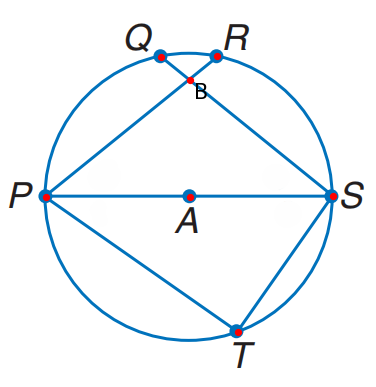}
\end{subfigure}

\begin{subfigure}[b]{0.24\linewidth}
    \centering
    \includegraphics[width=\linewidth]{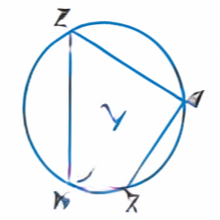}
    \caption{}
    \label{subfig:vae}
\end{subfigure}
\hfill
\begin{subfigure}[b]{0.24\linewidth}
    \centering
    \includegraphics[width=\linewidth]{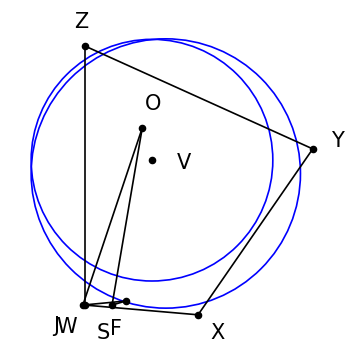}
    \caption{}
    \label{subfig:python}
\end{subfigure}
\hfill
\begin{subfigure}[b]{0.24\linewidth}
    \centering
    \includegraphics[width=\linewidth]{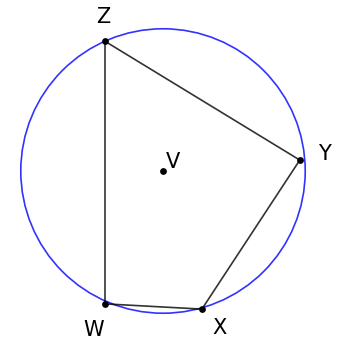}
    \caption{}
    \label{subfig:logic}
\end{subfigure}
\hfill
\begin{subfigure}[b]{0.24\linewidth}
    \centering
    \includegraphics[width=\linewidth]{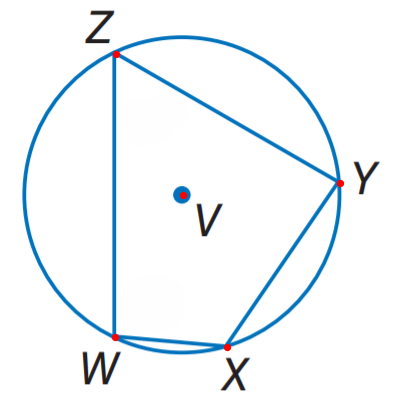}
    \caption{}
    \label{subfig:gt}
\end{subfigure}

\end{minipage}
} 

\vspace{-0.2cm}
\caption{
Reconstruction visualizations of input diagrams~\subref{subfig:gt} by
VQ-GAN~\subref{subfig:vae}, 
PyhParser~\subref{subfig:python}, 
\ourparser{}~\subref{subfig:logic}.
}
\label{fig:recons_vis}
\vspace{-0.5cm}
\end{figure}

\subsection{Ablation Study}
\label{exp:abla}
\noindent \textbf{Capturing Symbolic Understanding.}
To validate that structured representations are superior to feature vectors or code-based programming for diagram reconstruction, we compare three paradigms: (1) pixel-level auto-encoders (same scale diagram–fine-tuned VAE and VQ-GAN), (2) Python-code generation (PyhParser)—trained on the same samples as \ourparser{} but with logic forms translated t==o Python code, and (3) our \ourparser.  As shown in Tab.~\ref{table:recons}, symbolic representations achieve markedly higher reconstruction fidelity than pixel-level baselines. \ourparser-3B reduces MSE by 34.7\% (7.64 \vs 11.7) and improves DINO similarity by 14.8\% (0.93 \vs 0.81) compared to VQ-GAN. Visualizations (Fig.~\ref{fig:recons_vis}) further reveal that pixel-level methods lose fine structural details, while code-based models fail to capture symbolic semantics—generating points and lines without coherent shape or relational understanding. This limitation becomes more evident in diagram understanding tasks (Tab~\ref{table:perception}), where PyhParser attains only 11.4\% average accuracy, failing catastrophically on planar geometry (2.77\%) and graphs (0.0\%). \ourparser{} achieves 65.4\% accuracy—a 54.0\% absolute improvement. Manual inspection further confirms that Python-based parsers suffer from domain gaps and syntactic brittleness, whereas symbolic logic forms—being closer to natural language—enable effective transfer learning with minimal supervision.  

Other interesting findings include the model’s ability to transfer basic planar primitives (point vertices) to solid-geometry attribution, assigning colors and material properties (\eg, blue, metal). Moreover, after lightweight LoRA fine-tuning on a conversational diagram-perception dataset~\cite{sun2025mathglance} with \textit{direct-answer} supervision, the model can translate symbolic logic-form rules into linguistic \textit{Chain-of-Thoughts}, producing interpretable stepwise reasoning (\eg, Step~1: shape identification; Step~2: shape property analysis). Demos are provided in the Supp. Mat. (\S\ref{supp:expresults}).

 \noindent \textbf{Effect of Hierarchical Process Reward (HPR).}
We ablate three models to evaluate the effectiveness of HPR: \ourparser{}, \ourrl{}, and \ourrl(flat.). For reconstruction tasks (Tab.~\ref{table:recons}), hierarchical reward modeling (\ourrl{}) yields moderate gains, while achieving a larger +3.9\% improvement in diagram understanding, particularly on relation identification (\textit{rlat}) and object counting (\textit{cnt}) tasks (Tab.~\ref{table:perception}). The flat reward variant (\ourrl(flat.)), which averages all six reward components independently, performs comparably to the baseline, indicating that hierarchical dependencies are crucial for effective compositional learning.

\noindent \textbf{Stabilization Strategies for \ourvae.}
Achieving a self-supervised symbolic auto-encoder requires incorporating visual reconstruction rewards between reconstructed and ground-truth diagrams. However, we empirically observe that directly applying standard GRPO with perceptual losses leads to training collapse due to low reward variance—a critical issue for semantically sparse diagrams. As shown in Tab.~\ref{table:recons}, using visual rewards with vanilla GRPO ({\ourvae$\ominus$}) only yields only marginal improvement (0.65 in MSE, 0.02 in LPIPS). Training curves in Fig.~\ref{fig:method_fig} show reward scores quickly plateau and KL divergence remains near zero, indicating poor exploration in the policy space. In contrast, our stabilization strategies—hard negative contrastive learning ({\ourvae$\oplus$}) and power normalization annealing —yield steady reward growth and stable training. We adopt power normalization as the default setting due to its superior final performance and convergence stability.

\noindent \textbf{Neuro-symbolic System Designs.} 
Instead of integrating visual tokens into MLLMs—the conventional multimodal setting—we adopt a neuro-symbolic architecture where \ourrl{} converts diagrams into executable logic forms, which are prepended to the question and processed by an LLM reasoner (Qwen2.5-Math-7B). We evaluate this design on the GeoQA (Tab.~\ref{table:geoqa}). A visual reconstruction reward (Eq.~\ref{eq:vis}) is incorporated to complement textual answer accuracy (final-answer and format rewards), jointly guiding policy optimization. Empirically, low visual rewards often lead to malformed rollouts, whereas higher visual rewards promote coherent reasoning grounded in visual cues. To mitigate the influence of low-quality visual grounding, we introduce an adaptive reward weighting mechanism that dynamically scales the visual reward based on mathematical correctness—downweighting it when final answers are incorrect (Eq.~\ref{supp:eq:vis} in Supp. Mat.). \textit{Compared with direct visual-token integration}, the neuro-symbolic system offers explicit interpretability and modular supervision but remains less flexible in aligning symbolic visual representations with pretrained textual reasoning spaces. We hypothesize that reinforcement learning primarily refines existing reasoning policies rather than inducing new multimodal knowledge, aligned with the findings~\cite{yue2025does}.

\vspace{-.5em}
\section{Conclusion}
\vspace{-.5em}

We introduce the first symbolic auto-encoder that encodes diagrams into primitive-level latent spaces not pixel-level embeddings, supported by hierarchical process rewards and stabilized policy learning strategies. Experiments across reconstruction, perception, and reasoning tasks demonstrate its effectiveness and generalization from low-level to high-level diagram applications. Future research should extend symbolic auto-encoding to broader scientific and engineering domains, integrating verifiable/interpretable logical representations with dynamic world modeling, and developing reasoning-aligned visual supervision. 

{
    \small
    \bibliographystyle{unsrt}
    \bibliography{main}
}

\appendix

    
        
\setcounter{figure}{4}
\setcounter{table}{6}

\section{Synthetic Data Construction}
\label{supp:datasyn}
To train the symbolic vision encoder for diagram parsing, we develop a synthetic data generation pipeline that produces paired (diagram, logic form) examples. This pipeline uses computational geometry to generate diverse geometric configurations with accompanying structured symbolic representations. Algorithm~\ref{alg:data_generation} summarizes the complete synthetic data generation pipeline, integrating theorem construction, diagram rendering, logic form enrichment, and conversation formatting.

\subsection{Theorem Generation via Constructive Synthesis}

Our data generation process begins with synthetic theorem construction, following AlphaGeometry~\cite{trinh2024alphageometry}. Unlike AlphaGeometry, we generate only geometric configurations through a sequence of constructive operations, without requiring any deductive reasoning steps.

\noindent\textbf{Construction Operations.} We employ a fixed set of geometric construction primitives:
\begin{itemize}
\item \textbf{Point placement}: Random sampling in normalized coordinates $[0, 1]^2$
\item \textbf{Line construction}: Connect existing points or construct perpendiculars, parallels, and angle bisectors
\item \textbf{Circle construction}: Define circles by center and radius, or through three non-collinear points
\item \textbf{Intersection points}: Compute line-line, line-circle, and circle-circle intersections
\item \textbf{Derived constructions}: Midpoints, perpendicular bisectors, circumcenters, and incenters
\end{itemize}

\noindent\textbf{Configuration Sampling.} For each synthetic example, we randomly sample a construction sequence of length $k \sim \text{Uniform}(5, 15)$ operations. Each operation is selected with probability proportional to the current complexity budget, ensuring diverse configurations while avoiding degenerate cases (\eg, collinear points, overlapping circles). 

\noindent\textbf{Symbolic Representation.} Each configuration is serialized into a structured logic form $\mathcal{S}$ containing:
\begin{equation}
\mathcal{S} = \{\mathcal{P} \text{ (point names/positions)},\ 
\mathcal{E} \text{ (line instances)},\ 
\mathcal{G} \text{ (shape instances)},\ 
\mathcal{M} \text{ (shape indicators)},\ 
\mathcal{R} \text{ (geometric relations)}\},
\label{eq:logic_form_structure}
\end{equation}
where $\mathcal{P}$ encodes normalized coordinates, $\mathcal{E}$ represents connectivity, $\mathcal{G}$ describes shape types, $\mathcal{M}$ stores structural properties of shapes, and $\mathcal{R}$ captures primitive relations.

\subsection{Diagram Rendering from Logic Forms}

Given logic forms, we render a corresponding geometric diagram $\mathbf{I} \in \mathbb{R}^{512 \times 512 \times 3}$ using matplotlib's vector graphics engine.

\noindent\textbf{Rendering Configuration.} We enforce fixed plotting mode to ensure consistent visual appearance across the dataset:
\begin{itemize}
\item \textbf{Background}: White background ($\text{RGB} = (255, 255, 255)$)
\item \textbf{Point style}: Black scatter markers (size 15px)
\item \textbf{Line style}: Black solid lines (width 1.2px, opacity 0.8)
\item \textbf{Circle style}: Blue unfilled circles (width 1.2px, opacity 0.8)
\item \textbf{Label style}: Black text with automatic positioning (font size 15pt)
\end{itemize}

\noindent\textbf{Coordinate System.} Point coordinates in $\mathcal{P}$ are normalized to $[0, 1]^2$. We apply a $y$-axis flip transformation $y' = 1 - y$ to align with matplotlib's bottom-left origin convention. The viewport is dynamically scaled to encompass all geometric elements with a 40\% margin.


\subsection{Logic Form Enrichment}
\label{sec:logic_form_enrichment}

Raw synthetic configurations lack explicit structural properties that aid  diagram understanding. We augment each logic form with \textbf{indicator attributes} that describe shape-specific characteristics.
For each shape instance $g \in \mathcal{G}$, we use keyword matching to extract predefined spatial shape types:
\begin{itemize}[leftmargin=*]
  \item \textit{Triangles}: type (equilateral, isosceles, right) with angle and equality measurements.
  \item \textit{Quadrilaterals}: type (square, rectangle, parallelogram, trapezoid) with parallel/perpendicular relationships.
  \end{itemize}
These indicators are stored in the $\mathcal{M}$ component of the logic form (Eq.~\ref{eq:logic_form_structure}), providing rich semantic annotations for training.




\begin{algorithm}[t]
\caption{Synthetic Data Generation}
\label{alg:data_generation}
\begin{algorithmic}[1]
\Require Number of examples $N$, construction operation set $\Omega$
\Ensure Dataset $\mathcal{D} = \{(\mathbf{I}_i, \mathcal{S}_i, \mathcal{T}_i)\}_{i=1}^N$ where $\mathbf{I}$ is diagram, $\mathcal{S}$ is logic form, $\mathcal{T}$ is conversation

\For{$i = 1$ to $N$}
    \State \textcolor{gray}{// \textbf{Stage 1: Theorem Generation}}
    \State Sample construction sequence length $k \sim \text{Uniform}(5, 15)$
    \State Initialize empty configuration $\mathcal{Q} = \emptyset$
    \For{$j = 1$ to $k$}
        \State Sample operation $\omega \sim \Omega$ with complexity-weighted probability
        \State Apply $\omega$ to $\mathcal{Q}$ (add points, lines, circles, intersections)
        \If{configuration is degenerate (collinear points, overlapping circles)}
            \State Reject $\omega$ and resample
        \EndIf
    \EndFor
    \State Serialize $\mathcal{Q}$ to logic form $\mathcal{S}_i = \{\mathcal{P}, \mathcal{E}, \mathcal{G}, \mathcal{M}_{\text{empty}}, \mathcal{R}\}$

    \State
    \State \textcolor{gray}{// \textbf{Stage 2: Diagram Rendering}}
    \State Normalize point coordinates: $\mathcal{P} \gets \{(n, (x, 1-y)) \mid (n, (x, y)) \in \mathcal{P}\}$
    \State Compute viewport bounds $[x_{\min}, x_{\max}] \times [y_{\min}, y_{\max}]$ with 40\% margin
    \State Initialize canvas $\mathbf{I}_i \in \mathbb{R}^{512 \times 512 \times 3}$ with white background
    \State Render lines from $\mathcal{E}$ (black, width 1.2px)
    \State Render circles from $\mathcal{C}$ (blue, unfilled, width 1.2px)
    \State Render points and labels from $\mathcal{P}$ (black markers, font size 15pt)
    \State Save $\mathbf{I}_i$ as PNG file

    \State
    \State \textcolor{gray}{// \textbf{Stage 3: Logic Form Enrichment}}
    \For{each shape $g \in \mathcal{G}$}
      \State Extract shape type and constituent points
      \State Classify $g$ into special shape categories via keyword matching
      \State Create indicator $\iota_g \gets \{\text{shape}, \text{properties}\}$
        \State Add $\iota_g$ to indicator set $\mathcal{M}$
    \EndFor
    \State Update logic form: $\mathcal{S}_i \gets \{\mathcal{P}, \mathcal{E}, \mathcal{G}, \mathcal{M}, \mathcal{R}\}$

    \State
    \State \textcolor{gray}{// \textbf{Stage 4: Conversation Construction}}
    \State Create conversation $\mathcal{T}_i \gets \{\text{human}: \text{``\texttt{<image>} Generate logic forms $\cdots$''}, \text{gpt}: \mathcal{S}_i\}$

    \State Add $(\mathbf{I}_i, \mathcal{S}_i, \mathcal{T}_i)$ to dataset $\mathcal{D}$
\EndFor

\State Save diagrams $\{\mathbf{I}_i\}$ as PNG files, metadata $\mathcal{T}_i$ as JSON
\State \Return $\mathcal{D}$
\end{algorithmic}
\end{algorithm}

\begin{figure}[t]
    \centering
    \begin{subfigure}[b]{0.48\linewidth}
        \centering
        \includegraphics[width=\linewidth]{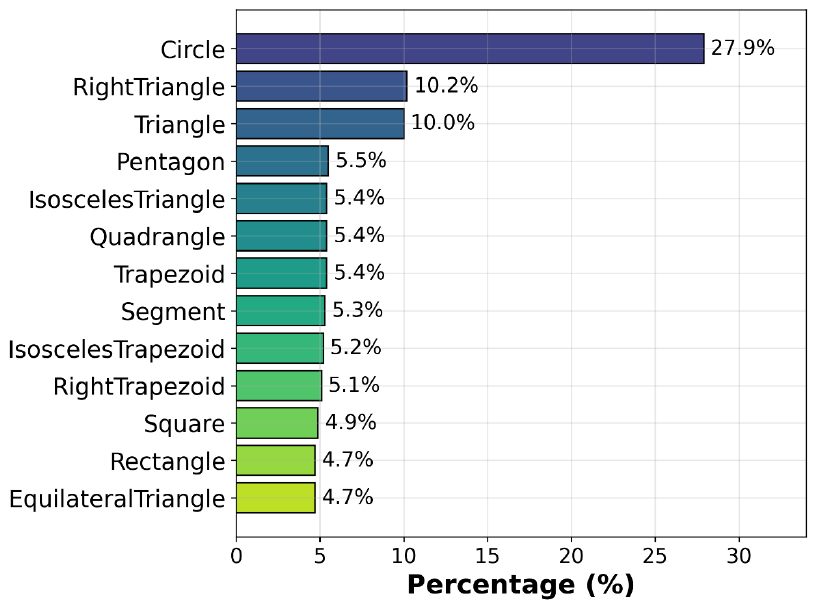}
        \caption{Shape distribution}
        \label{fig:shape_dist}
    \end{subfigure}
    \hfill
    \begin{subfigure}[b]{0.48\linewidth}
        \centering
        \includegraphics[width=\linewidth]{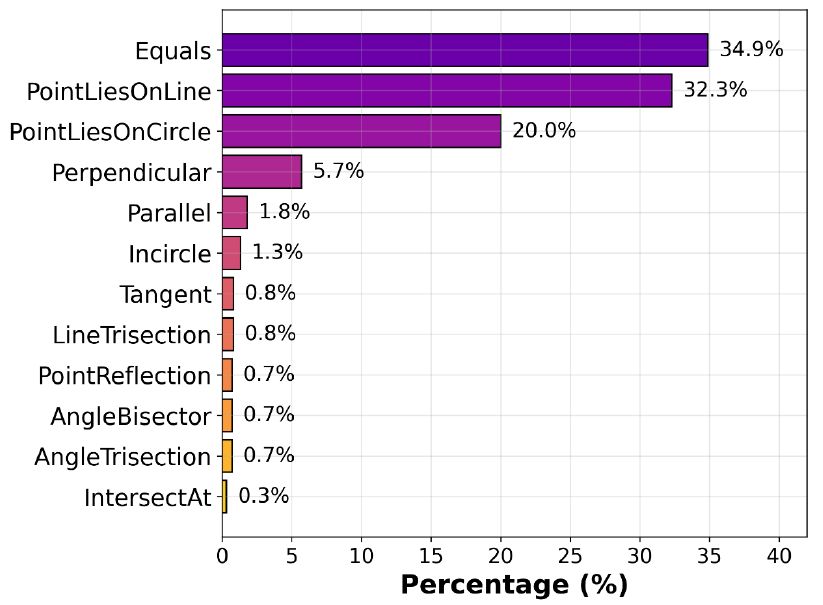}
        \vspace{-4mm}
        \caption{Relation distribution}
        \label{fig:rel_dist}
    \end{subfigure}
    \caption{
        Distributions of geometric shapes and relations in our synthetic dataset. The dataset exhibits diverse shape categories and relation types, providing  comprehensive supervision for learning topological diagram structures and hierarchical geometric relationships.
    }
    \label{fig:dataset_stats}
\end{figure}

\subsection{Conversation Dataset Construction}

To train the model in an instruction-following format, we convert logic forms into multi-turn conversational data.

\noindent\textbf{Conversation Template.} Each training example $({\mathbf{I}}, \mathcal{S})$ is formatted as:
\begin{verbatim}
{
  "id": "sample_*",
  "image": "/path/to/diagram.png",
  "conversations": [
    {
      "from": "human",
      "value": "<image>\nGenerate the logic forms
                for this geometric diagram."
    },
    {
      "from": "gpt",
      "value": "logic form"
    }
  ]
}
\end{verbatim}

\section{Data Statistics}
\label{supp:statis}


During training, we generate a 105K synthetic dataset, consisting of a 100K subset used to train \ourparser{} and a 5K subset used for \ourrl{}. On average, each diagram contains 6 points, 7 lines, 2 polygonal shapes, and 1 circle, reflecting a moderate level of geometric complexity. The shape distribution is well-balanced across 13 predefined categories, with \texttt{Circle} (27.9\%), \texttt{RightTriangle} (10.2\%), and \texttt{Triangle} (10.0\%) appearing most frequently, while the remaining categories each contribute between 4.7\% and 5.5\%. For shape properties, \texttt{Parallel} (55.5\%), \texttt{Perpendicular} (22.4\%), and \texttt{Equals} (22.1\%) relations collectively capture the dominant geometric constraints in typical constructions. At the relational level, \texttt{Equals} constraints constitute the largest proportion (34.9\%), followed by \texttt{PointLiesOnLine} (32.3\%) and \texttt{PointLiesOnCircle} (20.0\%). Less frequent but crucial relations—including \texttt{Perpendicular} (5.7\%), \texttt{Parallel} (1.8\%), \texttt{Incircle} (1.3\%), and \texttt{Tangent} (0.8\%)—ensure coverage of more intricate geometric configurations. This diverse composition provides a rich supervisory signal, enabling the model to acquire robust and hierarchical representations of geometric primitives, constraints, and structural dependencies. The full distributions of shape types and geometric relations are illustrated in Figs.~\ref{fig:shape_dist} and \ref{fig:rel_dist}.


\section{Logic Forms}
\label{supp:logicform}
\begin{figure}[t]
    \centering
    \includegraphics[width=\textwidth]{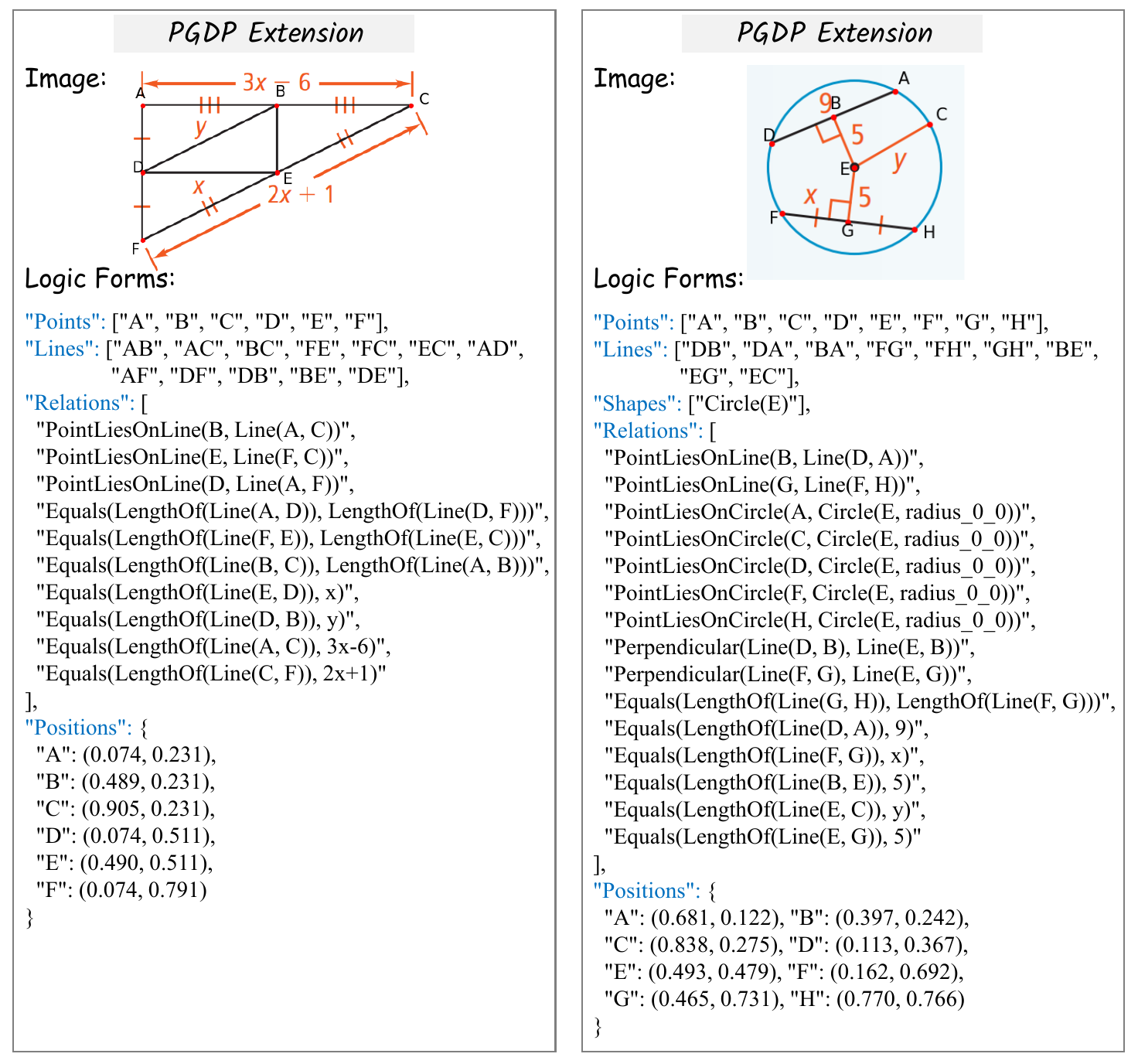}
    \caption{PGDP extension cases: original PGDP diagram alongside its enriched logic-form representation.}
   \label{fig:logicform_pgdp2}
\end{figure}

During hierarchical reward modeling, we use a 9K dataset comprising 5K synthetic samples and 4K diagrams from an extended version of PGDP~\cite{zhang2022pgdp}, where we augment the original diagrams with additional geometric relationships. Moreover, we corrected several annotation errors in the PGDP dataset. In particular, many diagrams lacked explicit vertex point annotations required for our logic-form representation. We manually added the missing vertex points to ensure complete geometric specifications—for example, points A–F in Fig.~\ref{fig:logicform_pgdp2} (left) and points A–G in Fig.~\ref{fig:logicform_pgdp2} (right).

\begin{figure}[t]
    \centering
    \includegraphics[width=\textwidth]{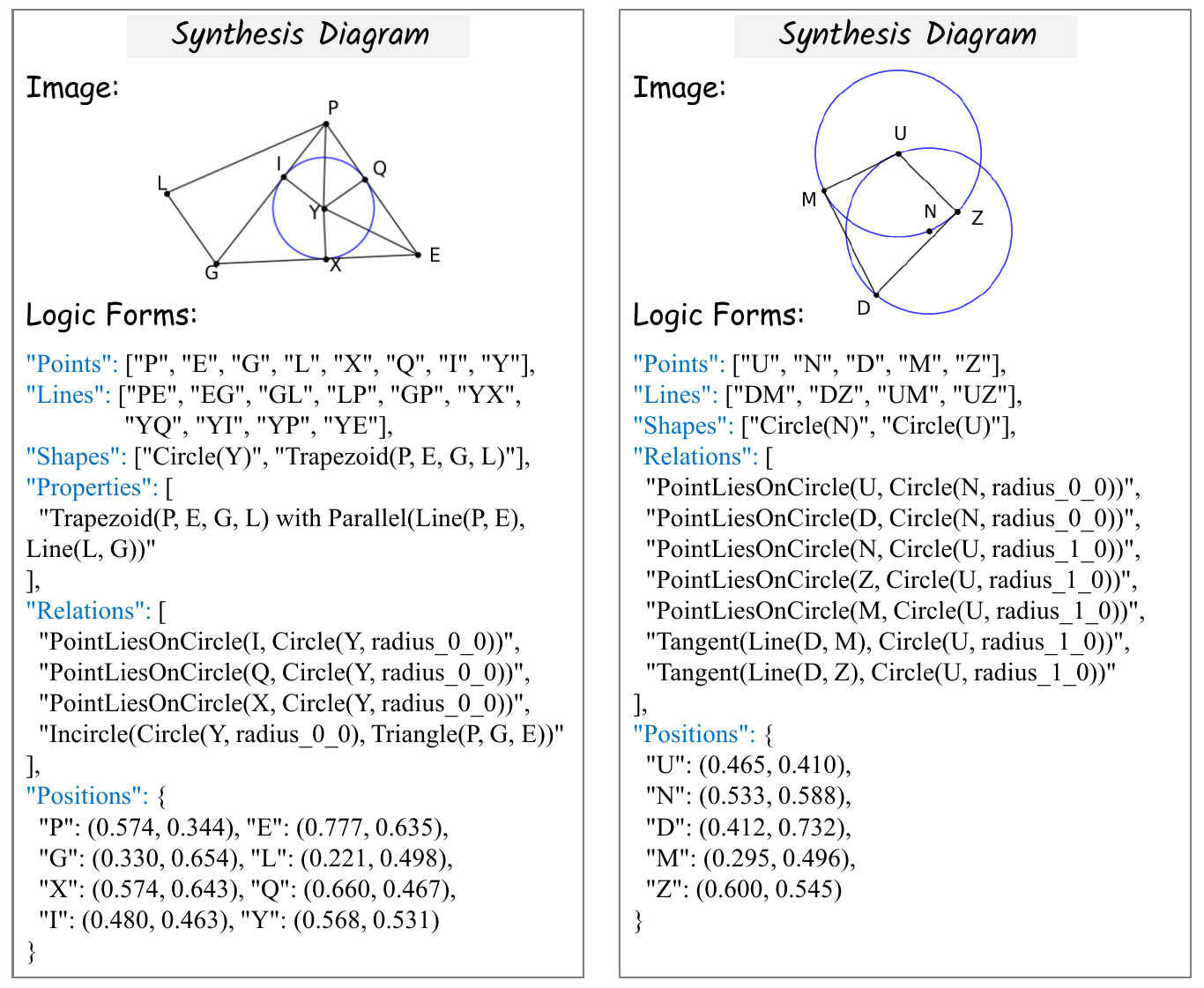}
    \caption{Visualizations of synthetic diagrams paired with their corresponding logic forms.}
   \label{fig:logicform_syn1}
\end{figure}

\begin{figure}[t]
    \centering
    \includegraphics[width=\textwidth]{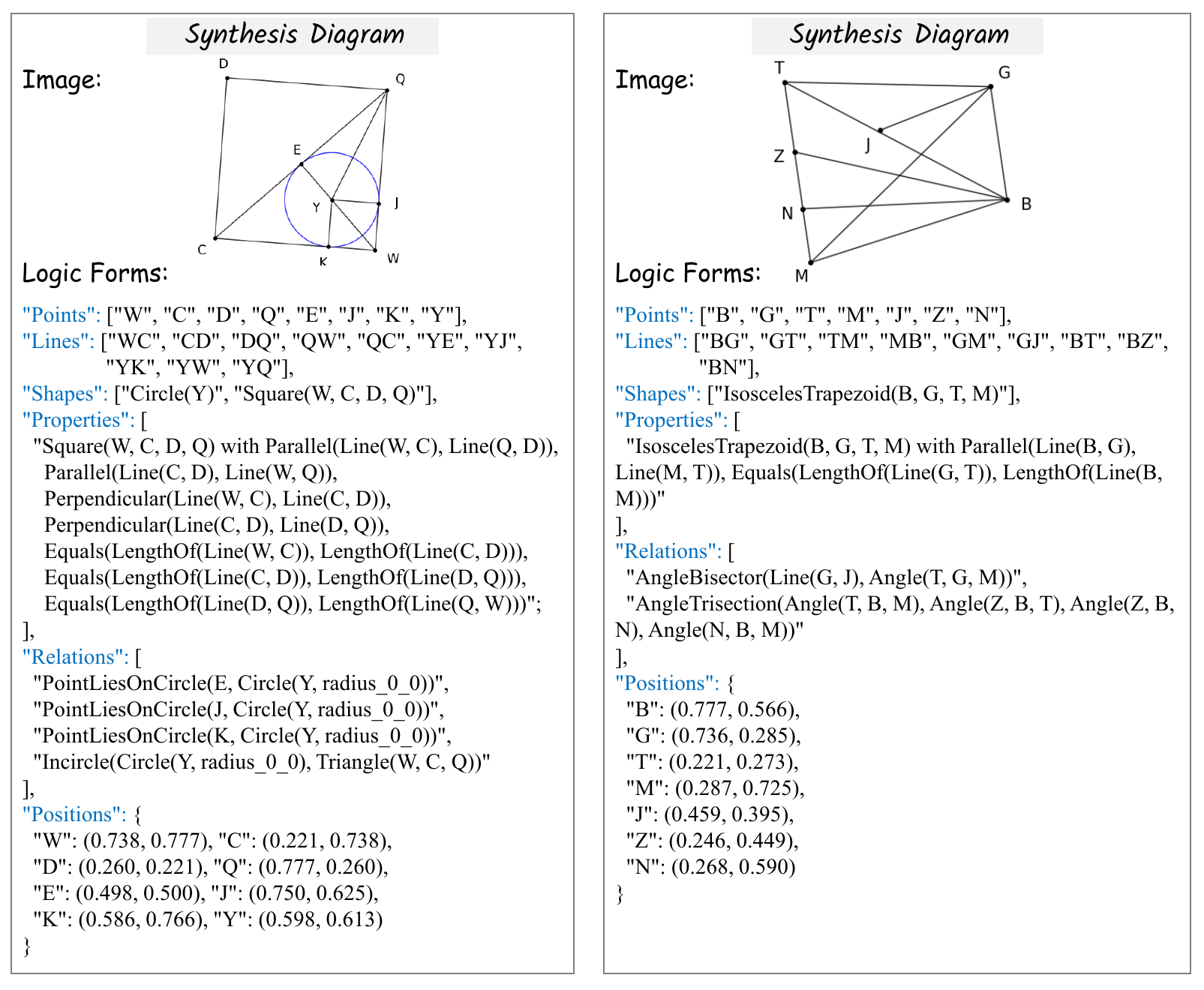}
    \caption{Visualizations of synthetic diagrams paired with their corresponding logic forms.}
   \label{fig:logicform_syn2}
\end{figure}

\begin{figure}[t]
    \centering
    \includegraphics[width=\textwidth]{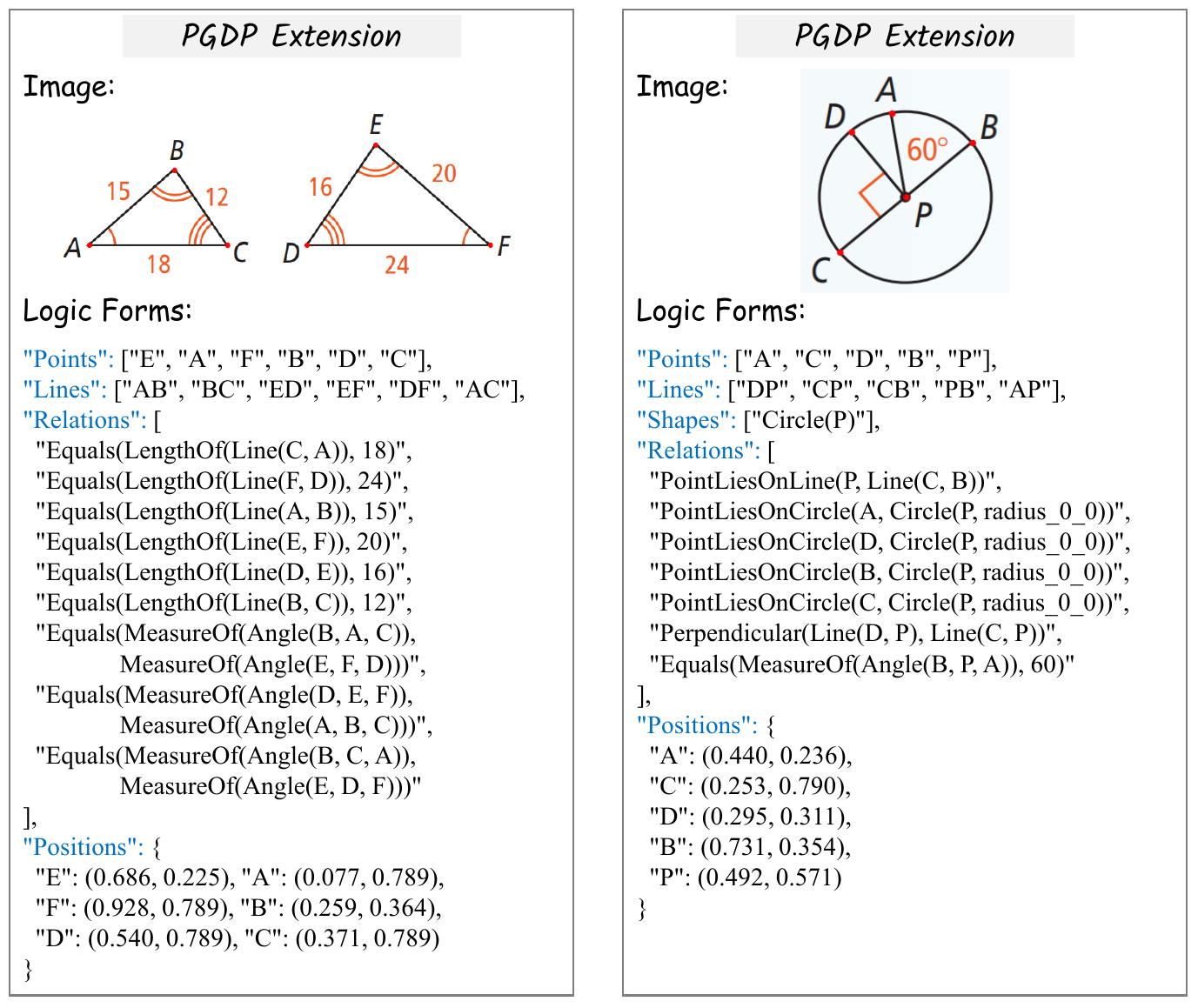}
    \caption{PGDP extension cases: original PGDP diagram alongside its enriched logic-form representation.}
   \label{fig:logicform_pgdp1}
\end{figure}

Figs.~\ref{fig:logicform_syn1}--\ref{fig:logicform_pgdp2} visualize the symbolic logic forms alongside their rendered diagrams. Each example illustrates the full hierarchical structure: normalized point coordinates, line segments connecting vertices, shape primitives with geometric indicators (\eg, parallel sides, perpendicular edges), and inter-primitive relations (\eg, \texttt{PointLiesOnLine}, \texttt{Incircle}).

\section{Symbolic Decoder}
\label{supp:decoder}
Our reconstruction decoder converts structured logic-form representations into geometric diagrams to compute visual similarity rewards in Eq.~\ref{eq:vis}. The decoder follows a multi-stage pipeline: it first parses the symbolic representations to extract geometric primitives (points, lines, and circles), computes derived attributes such as circle centers and radii, and finally renders the diagrams using computational geometry libraries (\eg, \texttt{matplotlib}). We summarize the entire process in Algorithm~\ref{alg:decoder}.
\subsection{Parsing Process}
\noindent\textbf{Point Positions.} We extract point/vertex coordinates using regular expression matching:
\begin{equation}
\mathcal{P} = \{(n_i, \mathbf{p}_i) \mid i = 1, \ldots, N_p\},
\end{equation}
where $n_i$ denotes the point name and $\mathbf{p}_i = (x_i, y_i) \in [0, 1]^2$ represents normalized coordinates. We apply a coordinate transformation $y'_i = 1 - y_i$ to align with the bottom-left origin convention.

\noindent\textbf{Line Instances.} Line segments are represented as ordered pairs of point names:
\begin{equation}
\mathcal{E} = \{(n_j^{(1)}, n_j^{(2)}) \mid j = 1, \ldots, N_\ell\}
\end{equation}
Each line instance string (\eg, ``ab'') is decomposed into constituent point identifiers that index into $\mathcal{P}$.

\noindent\textbf{Circle Instances.} Circles are identified through two mechanisms: (1) explicit declarations in shape representations (\eg, \texttt{Circle(o)}), extracting center point names, and (2) implicit definitions via geometric relations. We maintain a set of circle center identifiers:
\begin{equation}
\mathcal{C} = \{c_k \mid k = 1, \ldots, N_c\}
\end{equation}
For the implicit geometric relations, we parse three types of circle-defining relations that enable radius computation, as follows:

\noindent (a) From \texttt{PointLiesOnCircle(p, Circle(c, radius))} relations, we compute the radius as the Euclidean distance:
\begin{equation}
r_c = \|\mathbf{p}_p - \mathbf{p}_c\|_2
\label{eq:point_circle}
\end{equation}
(b) For \texttt{ConcyclicPoints($p_1, p_2, p_3, p_4$)}, we fit a circumcircle using least-squares optimization. The circle equation $(x - a)^2 + (y - b)^2 = r^2$ is linearized to:
\begin{equation}
x^2 + y^2 = 2ax + 2by + c, \quad \text{where } c = r^2 - a^2 - b^2
\end{equation}
Given $n \geq 3$ points, we solve the overdetermined linear system:
\begin{equation}
\begin{bmatrix} x_1 & y_1 & 1 \\ \vdots & \vdots & \vdots \\ x_n & y_n & 1 \end{bmatrix} \begin{bmatrix} 2a \\ 2b \\ c \end{bmatrix} = \begin{bmatrix} x_1^2 + y_1^2 \\ \vdots \\ x_n^2 + y_n^2 \end{bmatrix}
\label{eq:circumcircle_lstsq}
\end{equation}
via least-squares to obtain center $\mathbf{c} = (a, b)$ and radius $r = \sqrt{c + a^2 + b^2}$. We reject solutions where the maximum fitting error exceeds 10\% of the radius.

\noindent (c) From \texttt{Incircle(Circle(c, radius), Triangle($v_1, v_2, v_3$))}, we compute the incircle radius using Heron's formula. For triangle side lengths $a, b, c$ and semiperimeter $s = (a + b + c)/2$:
\begin{equation}
r = \frac{A}{s}, \quad \text{where } A = \sqrt{s(s-a)(s-b)(s-c)}
\label{eq:incircle_heron}
\end{equation}

\subsection{Rendering Pipeline}
\label{sec:decoder_rendering}

The rendering process transforms the parsed geometric primitives into a rasterized image $\mathbf{I} \in \mathbb{R}^{H \times W \times 3}$. For visual similarity computation, the reconstructed diagrams are resized to match the ground-truth sizes, ensuring identical resolution for alignment.

\noindent\textbf{Coordinate System Setup.} We compute the axis-aligned bounding box encompassing all points and circles:
\begin{align}
x_{\min} &= \min_{i} \{x_i\} - \min_k \{r_k\}, \quad x_{\max} = \max_{i} \{x_i\} + \max_k \{r_k\} \label{eq:bbox_x} \\
y_{\min} &= \min_{i} \{y_i\} - \min_k \{r_k\}, \quad y_{\max} = \max_{i} \{y_i\} + \max_k \{r_k\} \label{eq:bbox_y}
\end{align}
A margin factor $\alpha = 0.4$ is applied to extend the viewport:
\begin{equation}
[x_{\min}, x_{\max}] \leftarrow [x_{\min} - \alpha \Delta_x, x_{\max} + \alpha \Delta_x]
\label{eq:viewport_margin}
\end{equation}
where $\Delta_x = x_{\max} - x_{\min}$, and similarly for the $y$-axis. Equal aspect ratio is enforced to prevent distortion.

\noindent\textbf{Primitive Rendering.} Geometric elements are rendered as follows:
(a) \textit{Lines}: Each line $(n_j^{(1)}, n_j^{(2)}) \in \mathcal{E}$ is rasterized as a black line segment connecting $\mathbf{p}_{j}^{(1)}$ and $\mathbf{p}_{j}^{(2)}$ with linewidth 1.2 pixels and opacity $\alpha = 0.8$; (b) \textit{Circles}: Each circle $c_k \in \mathcal{C}$ is rendered as an unfilled circle with center $\mathbf{p}_{c_k}$ and radius $r_k$, using blue stroke color (RGB: 0, 0, 255), linewidth 1.2 pixels, and opacity $\alpha = 0.8$; (c) \textit{Points and Names}: Each point $\mathbf{p}_i$ is drawn as a black scatter marker (size 15 pixels, z-order 10). Point names are positioned using the naming algorithm~\cite{trinh2024alphageometry}, which computes smart offsets to avoid overlaps with lines and circles.  The rendered canvas is saved as a PNG image with white background at 100 DPI.





\noindent\textbf{Failure Handling.} In cases where parsing fails or produces degenerate geometry (\eg, empty point sets, singular circle fits), the decoder generates a black image to signal reconstruction failure, resulting in minimal visual similarity reward.

\begin{algorithm}[t]
\caption{Diagram Reconstruction}
\label{alg:decoder}
\begin{algorithmic}[1]
\Require Logic forms $\mathcal{\hat{S}}$
\Ensure Rendered image $\mathbf{I} \in \mathbb{R}^{H \times W \times 3}$

\State Parse $\mathcal{\hat{S}}$ to extract:
\State \quad - Point set $\mathcal{P} = \{(n_i, \mathbf{p}_i)\}$ via regex matching
\State \quad - Line set $\mathcal{E} = \{(n_j^{(1)}, n_j^{(2)})\}$ from line instances
\State \quad - Circle set $\mathcal{C} = \{c_k\}$, derived from both explicit shape representations and implicit geometric relations.
\State Compute circle radii $\{r_k\}$ from geometric relations:
\State \quad - \texttt{PointLiesOnCircle}: $r_c \gets \|\mathbf{p}_p - \mathbf{p}_c\|_2$ (Eq.~\ref{eq:point_circle})
\State \quad - \texttt{ConcyclicPoints}: $(\mathbf{c}, r) \gets \textsc{FitCircleLeastSquares}(\{\mathbf{p}_1,\ldots,\mathbf{p}_4\})$ (Eq.~\ref{eq:circumcircle_lstsq})
\State \quad - \texttt{Incircle}: $r \gets \textsc{ComputeIncircleRadius}(\text{vertices})$ (Eq.~\ref{eq:incircle_heron})

\State Establish coordinate system:
\State \quad - Compute bounding box $[x_{\min}, x_{\max}] \times [y_{\min}, y_{\max}]$ via Eq.~\ref{eq:bbox_x}-\ref{eq:bbox_y}
\State \quad - Extend by margin factor $\alpha = 0.4$ via Eq.~\ref{eq:viewport_margin}
\State \quad - Set equal aspect ratio

\State Render primitives in order:
\State \quad - Draw points and names: $\forall (n_i, \mathbf{p}_i) \in \mathcal{P}$
\State \quad - Draw lines: $\forall (n_j^{(1)}, n_j^{(2)}) \in \mathcal{E}$
\State \quad - Draw circles: $\forall c_k \in \mathcal{C}$ with radius $r_k$

\State Save canvas as PNG $\to \mathbf{I}$
\State \Return $\mathbf{I}$
\end{algorithmic}
\end{algorithm}



\section{Training Details}
\label{supp:traindetail}
\noindent\textbf{Cold Start Training (\ourparser{}):} We initialize the symbolic vision encoder using supervised fine-tuning on a synthetic dataset of 100K diagram-logic form pairs, training for one epoch to establish foundational symbolic parsing capability. During cold start training, we use a global batch size of 256, with the learning rate to $1 \times 10^{-5}$ (a vision-specific learning rate of $2 \times 10^{-6}$). We employ the Adam optimizer without weight decay and apply a cosine learning rate schedule. To improve memory efficiency, we utilize Fully Sharded Data Parallel (FSDP), gradient checkpointing, and enable BF16 precision, avoiding CPU/GPU offloading to maximize throughput. All 3B/7B model training is performed on 8 $\times$ A100 GPUs (80GB each), with alignment training taking approximately 9-12 hours.

\noindent\textbf{Hierarchical Reward Modeling (\ourrl{}):} Hierarchical reward modeling uses reinforcement learning (RL) to capture the 
inherent structural hierarchy of geometric diagrams. We train the encoder on 
9{,}000 samples from our synthetic dataset and the extended PGDP data, enabling 
the model to learn compositional dependencies across levels: points $\Rightarrow$ lines $\Rightarrow$ shapes $\Rightarrow$ relations. We adopt the Verifiers framework~\cite{verifiers} and optimize using Group 
Relative Policy Optimization (GRPO). Our cluster configuration consists of 
3 nodes, with 1 inference node and 2 training nodes, each equipped with 
$2\times$~A100 GPUs (40GB) and 12 CPU cores, with a SLURM time limit of 2 days. 
We use a per-device batch size of 4 and generate 8 rollouts per prompt 
(GRPO group size). Combined with gradient accumulation over 2 steps, this yields 
an effective batch size of 32 unique prompts per update 
($4 \text{ prompts} \times 4 \text{ GPUs} \times 2 \text{ accumulation steps}$). 
The learning rate is set to $5 \times 10^{-7}$ with a KL penalty coefficient 
$\beta = 0.03$. The maximum generation length is 1{,}024 tokens.

\noindent\textbf{Self-Supervised Learning (\ourvae{}):} We train \ourvae{} on 16{,}000 diagrams, consisting of 5{,}000  samples from 
Geo170K~\cite{gao2023gllava}, 7{,}000 synthetic diagrams, and 4{,}000 diagrams from PGDP~\cite{zhang2022pgdp}. The model learns self-supervised reconstruction 
through perceptual rewards, enabling structure-aware diagram encoding. 
To stabilize RL optimization, we apply power normalization and noise annealing 
as described in the main paper (Sec.~\ref{main:stabel}).

\noindent\textbf{Downstream Task Fine-Tuning (\ourvae{}+$\ast$):} For downstream applications, we apply LoRA~\cite{hu2021lora} adaptation (rank=64) and train for one epoch on task-specific datasets. (1) \textbf{Perception (MathGlance ~\cite{sun2025mathglance}):} We fine-tune on 10K planar geometry diagrams (1/10 of the full 100K dataset) with direct-answer supervision to allow the model apot to the converation insttucitons and enhance primitive recognition capabilities;  (2) \textbf{Chart Reconstruction (VisCodex~\cite{Jiang2025VisCodex}):} We train on 100K chart-code pairs (1/5 of the full 598K dataset)to evaluate cross-domain symbolic representation transfer via chart-to-code generation; (3) \textbf{Reasoning:} We train on Geo170K~\cite{gao2023gllava} and MathV360K~\cite{shi2024mathllava} using question-answering with chain-of-thought supervision to enable mathematical problem reasoning tasks.

\noindent\textbf{Reconstructions Beyond Geometry:}
To assess the generality of our reconstruction framework beyond planar geometry, we conduct experiments on two out-of-domain diagram types: electrical circuits and molecular structures. Although circuit and molecular diagrams differ from geometric figures, a portion of their structure is still formed by basic geometric primitives such as points, lines, circles, and polygonal contours. This structural overlap enables our representation to transfer across domains. 
For circuits, we collect 2{,}903 images from the \texttt{electronics\_diagrams\_single} subset of MMMU-Pro~\cite{yue2025mmmupro} and EEE-Bench~\cite{li2025eeebench}. For molecular diagrams, we use 923 images from the MMChemOCR subset introduced in ChemVLM~\cite{li2025chemvlm}. To ensure that the extended domains remain compatible with our symbolic rules,  we adopt a preprocessing step that filters out samples with complex topologies, \eg, sinusoidal AC waveforms in circuits and wedge-dash notation for three-dimensional stereochemistry, based on the visual reconstruction reward (Eq.~\ref{eq:vis}). Specifically, we run \ourvae{} on all circuit and molecular images and retain only those examples with a visual reward greater than $0.6$ ($r_{\text{vis}} > 0.6$). After filtering, we obtain 536 circuit diagrams and 447 molecular diagrams for training. We use the same training configuration as in \ourvae{} for this experiment.

\section{\ourvae{} Reconstruction Visualizations}
\label{supp:recons}
  
\begin{figure}[t]
\vspace{-0.3cm}
    \centering
    \includegraphics[width=0.95\textwidth]{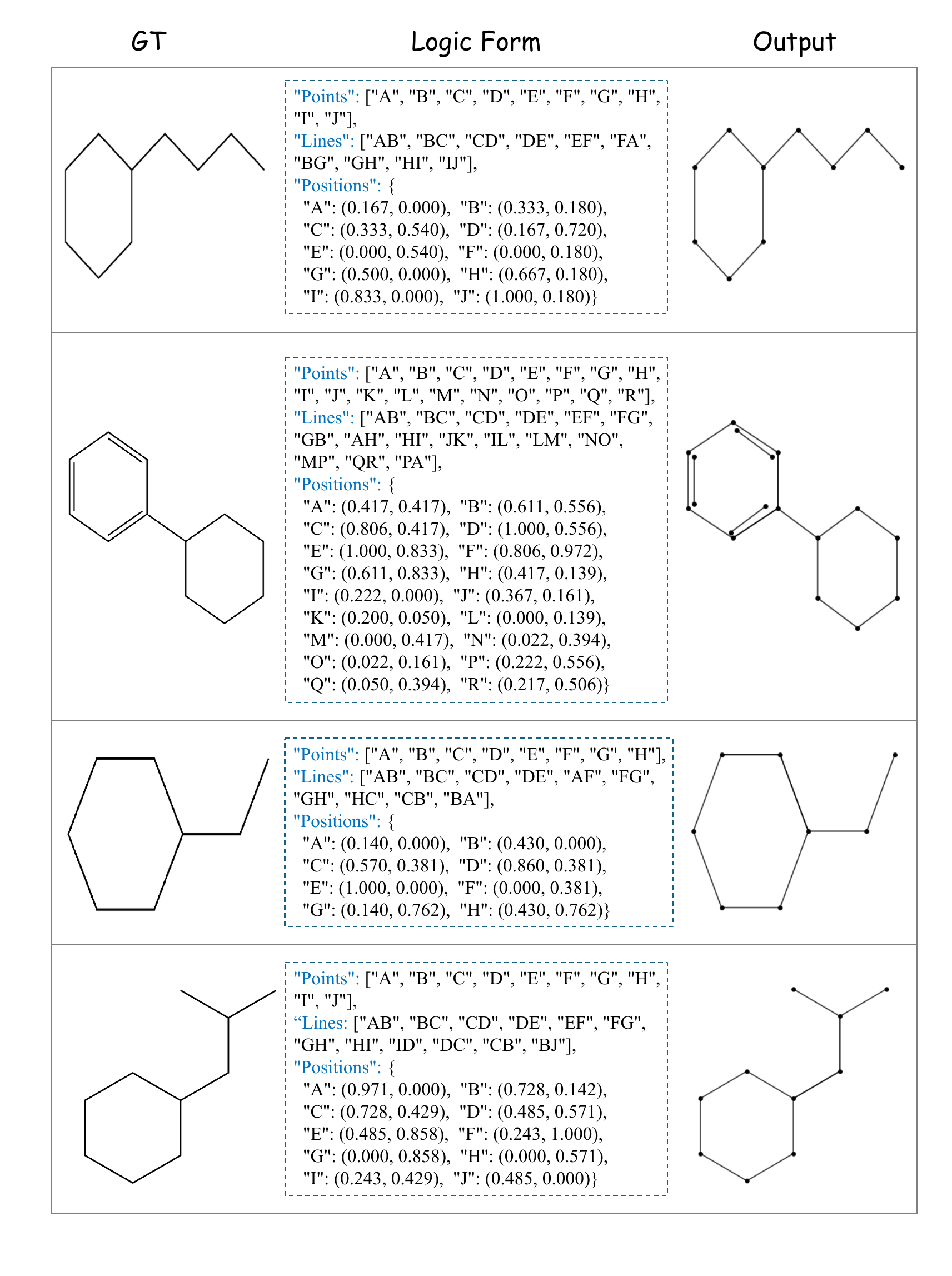}
    \caption{Reconstruction results of \ourvae{} on molecular structure diagrams. Each triplet shows the ground-truth input, the predicted logic-form primitives, and the corresponding reconstructed diagram.}
   \label{fig:corr_mol}
\end{figure}

\begin{figure}[t]
    \centering
    \includegraphics[width=0.95\textwidth]{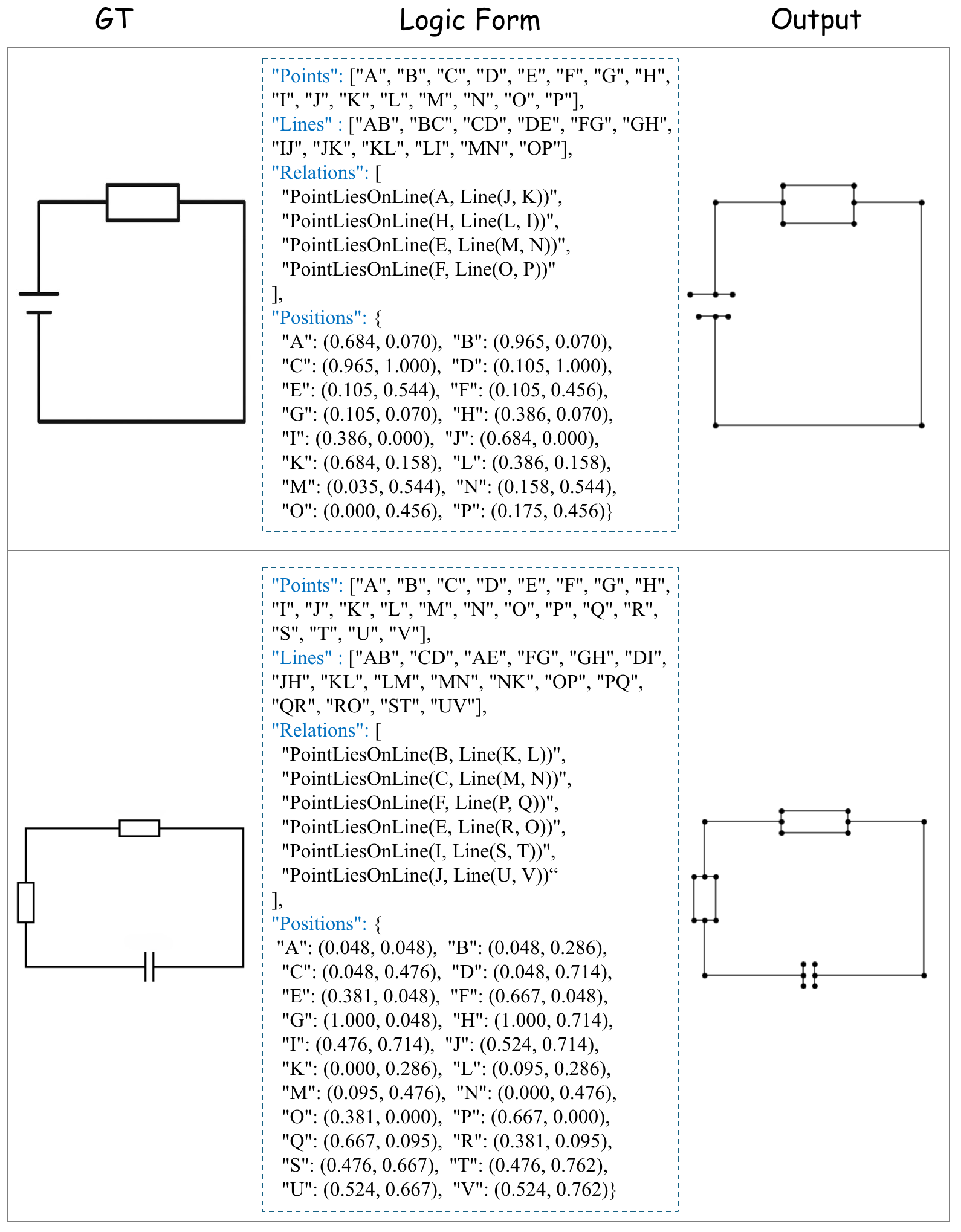}
    \caption{Reconstruction results of \ourvae{} on electrical circuit diagrams. Each triplet shows the ground-truth input, the predicted logic-form primitives, and the corresponding reconstructed diagram.}
   \label{fig:corr_cir}
\end{figure}

We provide reconstruction visualizations produced by \ourvae{} across electrical circuit diagrams and molecular structure diagrams in Figs.~\ref{fig:corr_mol}--\ref{fig:corr_cir}. These results demonstrate the model’s ability to recover both low-level primitives and global structural organization. In each demonstration, we present triplets consisting of the ground-truth input, the latent logic-form primitives, and the final reconstruction. For clarity, we omit vertex indicators in the visualizations. Since those circuit and molecular diagrams are primarily composed of basic geometric primitives, \eg, points, lines, and polygonal shapes, our model can faithfully reconstruct their topologies.

\begin{figure}[t]
    \centering
    \includegraphics[width=0.83\textwidth]{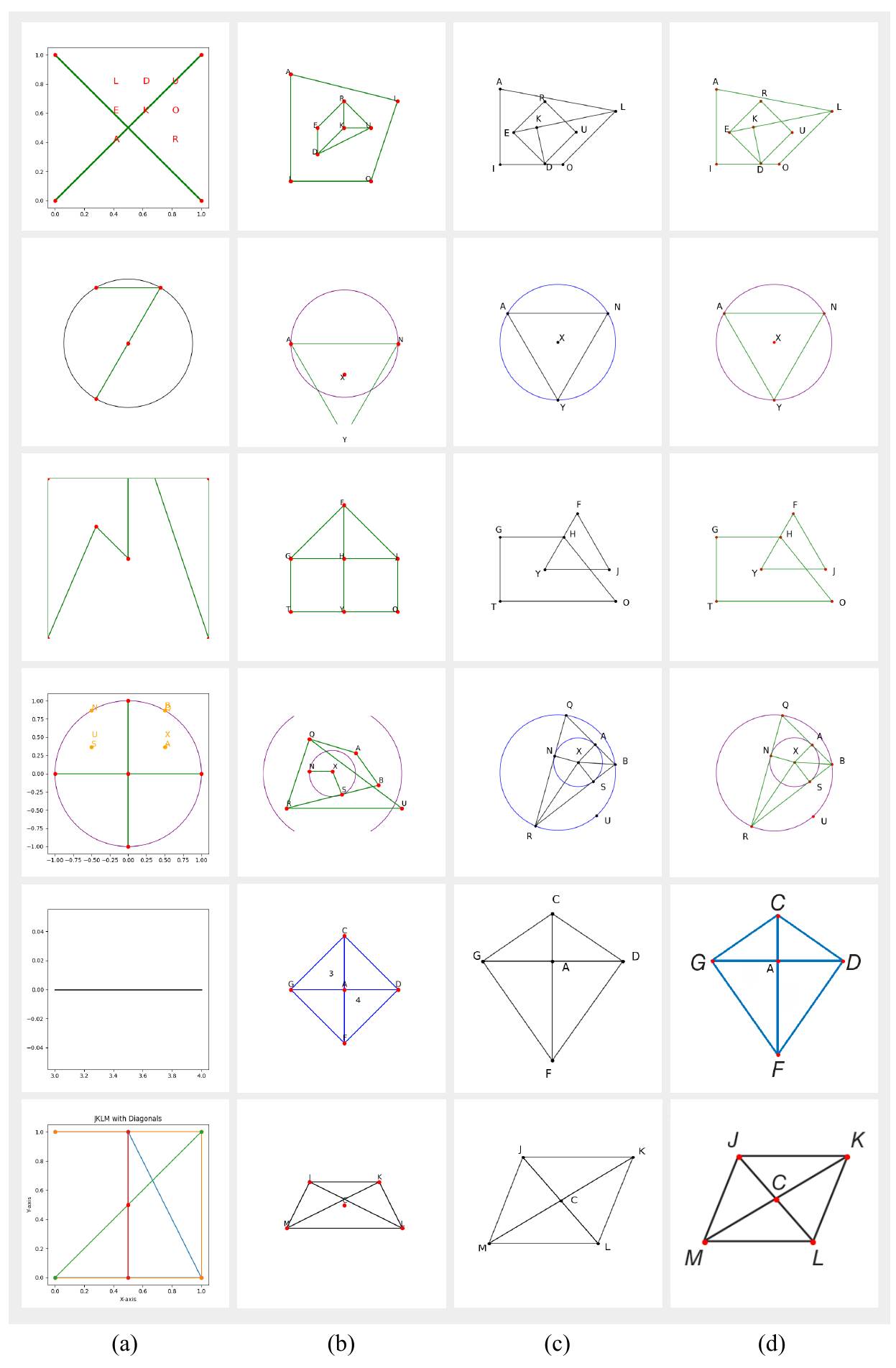}
    \caption{Qualitative visualizations on geomatic diagrams reconstruction \wrt Qwen2.5-VL-7B, GPT-4o, and \ourvae{}-7B. The final column shows the ground-truth inputs.}
   \label{fig:vis_compare}
\end{figure}

We further compare our geometric reconstructions with those generated by Qwen2.5-VL-7B (Fig.~\ref{fig:vis_compare} (a)) and GPT-4o (Fig.~\ref{fig:vis_compare} (b)). General-purpose multimodal models struggle to preserve fine-grained geometric structure, particularly in complex diagrams, whereas our approach yields substantially more faithful geometric and relational reconstructions (Fig.~\ref{fig:vis_compare} (c)). Notably, for these comparators, we prompt the models to generate Python plotting code and render the diagrams via Matplotlib, while our method reconstructs diagrams directly from symbolic logic forms through our dedicated decoder.

\begin{figure}[t]
    \centering
    \includegraphics[width=\textwidth]{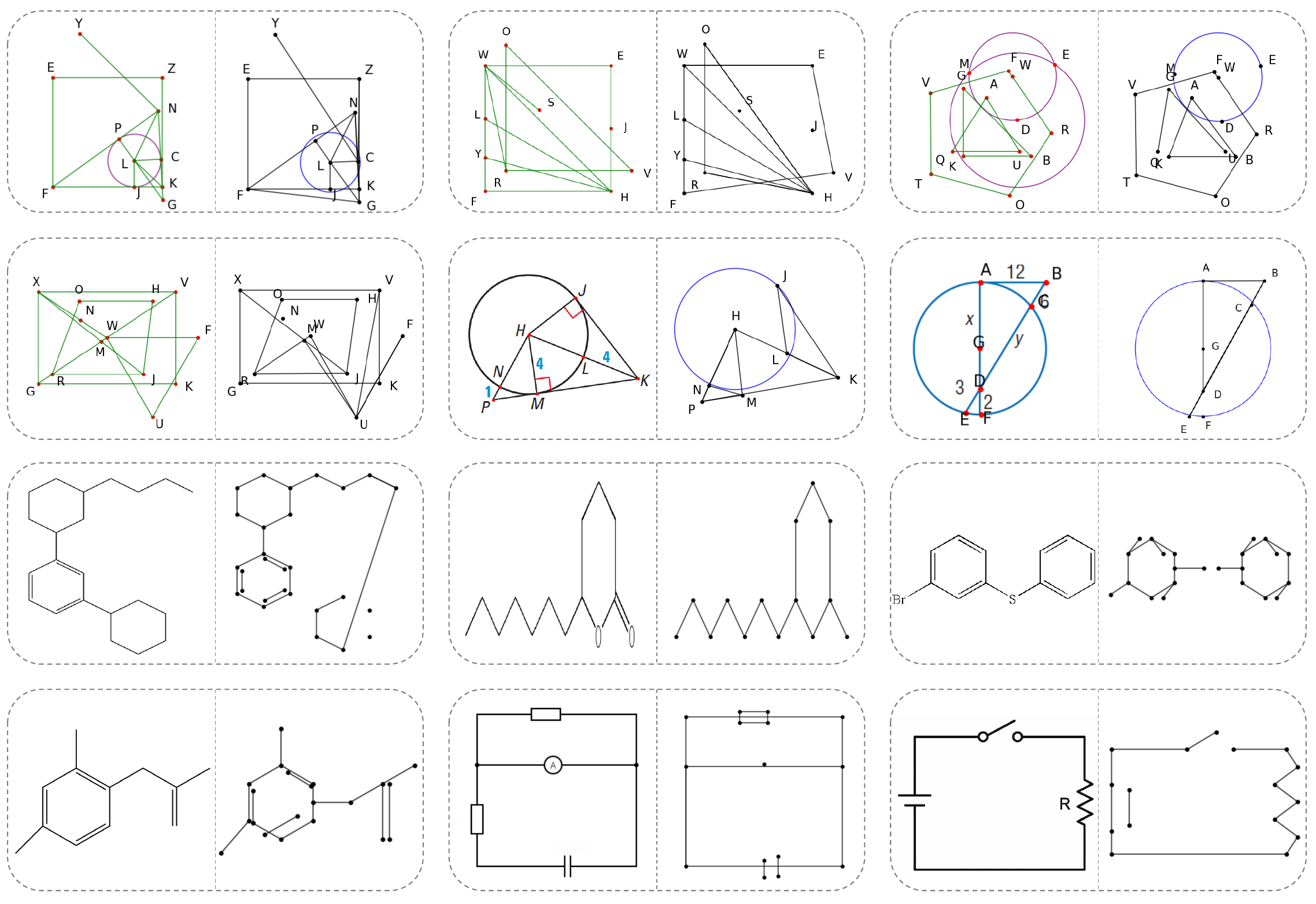}
    \caption{Visualizations of failure reconstructions produced by \ourvae{}.}
   \label{fig:fail_mol_cir}
\end{figure}

As discussed in Limitation~\ref{supp:limitation}, the rule-based nature of our symbolic design limits flexibility when extending the system to new geometric concepts. Consequently, when input diagrams contain structures outside our geometric primitives—particularly those with higher visual complexity or domain-specific topology—our model fails to fully reconstruct them. In Fig.~\ref{fig:fail_mol_cir}, the model is unable to recover domain-specific graphical symbols whose semantics lie beyond basic geometry. Examples include the wavy-line representation of resistors and the lettered ammeter symbol (circled ``A''). Moreover, for diagrams with dense geometric content, such as the clustered shapes shown in the top rows, our model struggles to recover all structural details accurately. We discuss potential extensions to address these limitations in further work (Sec.~\ref{supp:limitation}).

\section{Cross-Lingual and Cross-Modal Translation}
\label{supp:gene}

\noindent \textbf{Cross-Lingual Translation.} After fine-tuning on downstream perception tasks using the GeoPeP conversational-format samples, we observe an emergent cross-lingual transfer phenomenon during inference: formal logic forms are automatically translated into fluent natural-language chain-of-thoughts. This indicates that the model internalizes the symbolic structure encoded in our logic-form representation and can project it into coherent linguistic reasoning. Remarkably, this ability emerges without any explicit CoT supervision, suggesting that our logic forms are intrinsically aligned with natural-language semantics—an alignment that other program-like syntaxes (\eg, Python) fail to achieve.

\begin{figure}[t]
    \centering
    \begin{subfigure}{.95\textwidth}
        \centering
        \includegraphics[width=\textwidth]{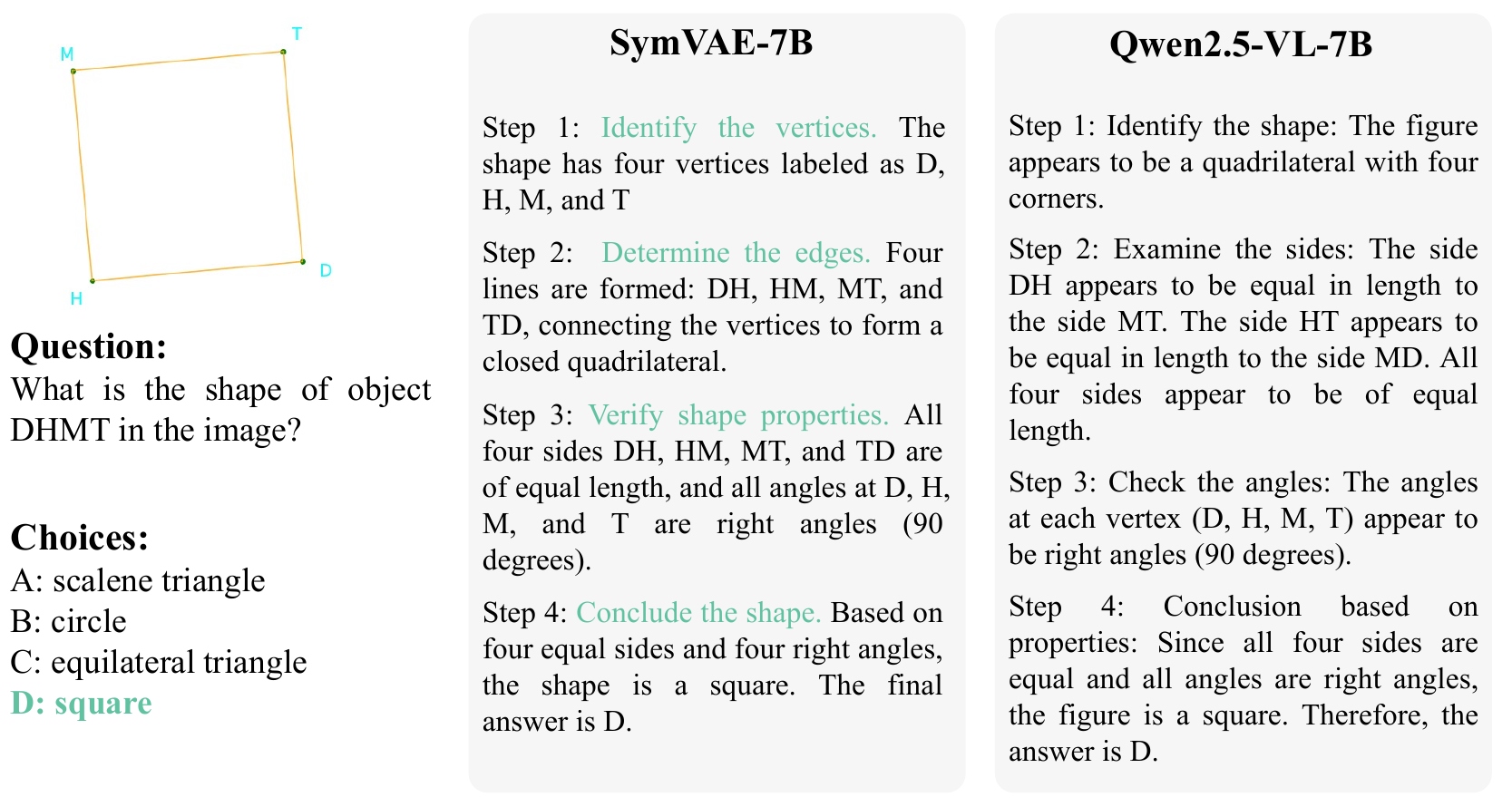}
        \caption{}
        \label{fig:cot_demo1}
    \end{subfigure}
    \hfill
    \begin{subfigure}{.95\textwidth}
        \centering
        \includegraphics[width=\textwidth]{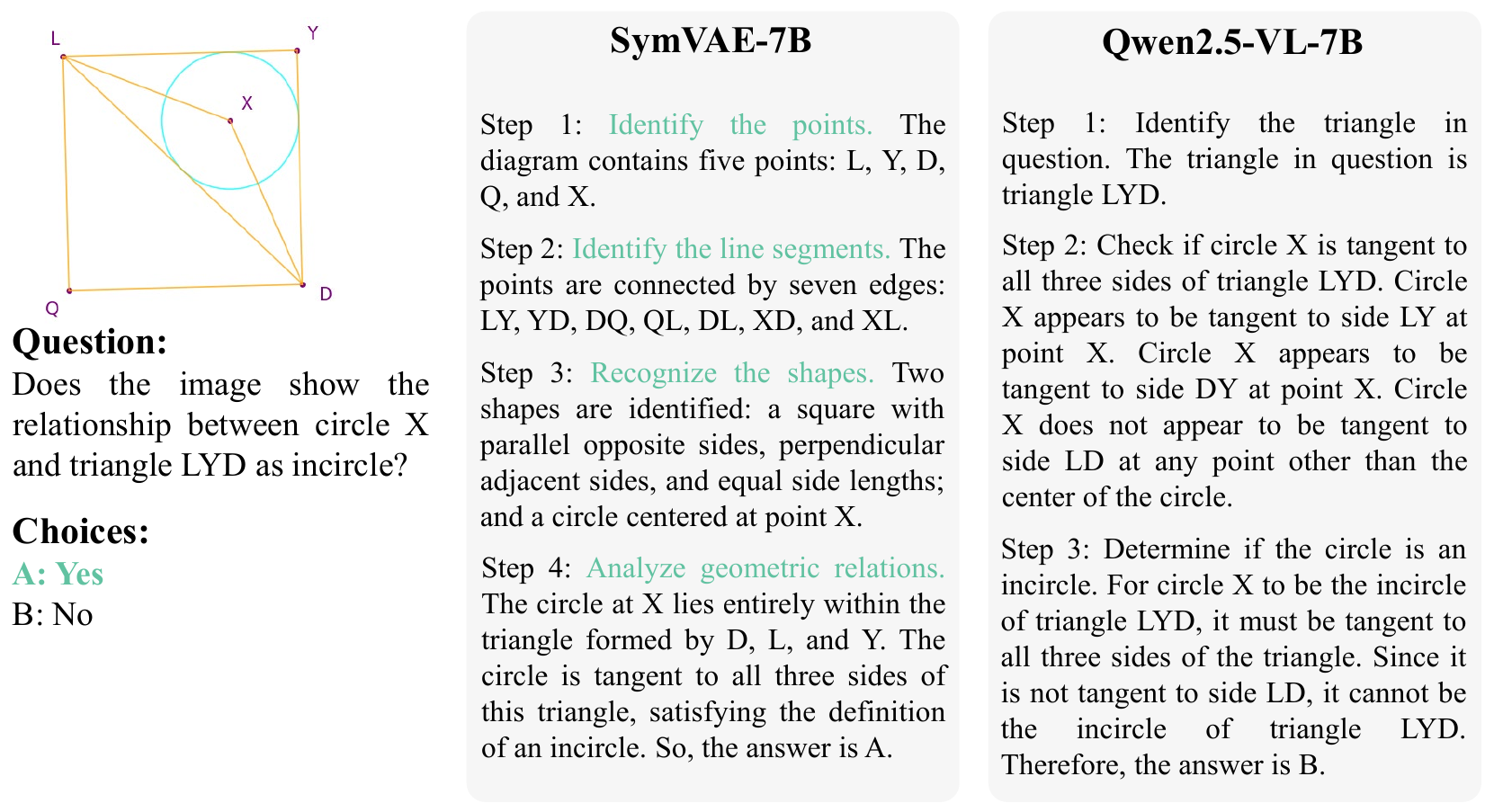}
        \caption{}
        \label{fig:cot_demo2}
    \end{subfigure}
    \caption{Emergent cross-lingual translation on MathGlance evaluation samples. The translated chain-of-thoughts are not simple mirrors of the priors embedded in the base model (Qwen2.5-VL-7B); instead, they follow the structured reasoning induced by our logic-form representation.}
    \label{fig:cot_comparison1}
\end{figure}

\begin{figure}[t]
    \centering
    \begin{subfigure}{.95\textwidth}
        \centering
        \includegraphics[width=\textwidth]{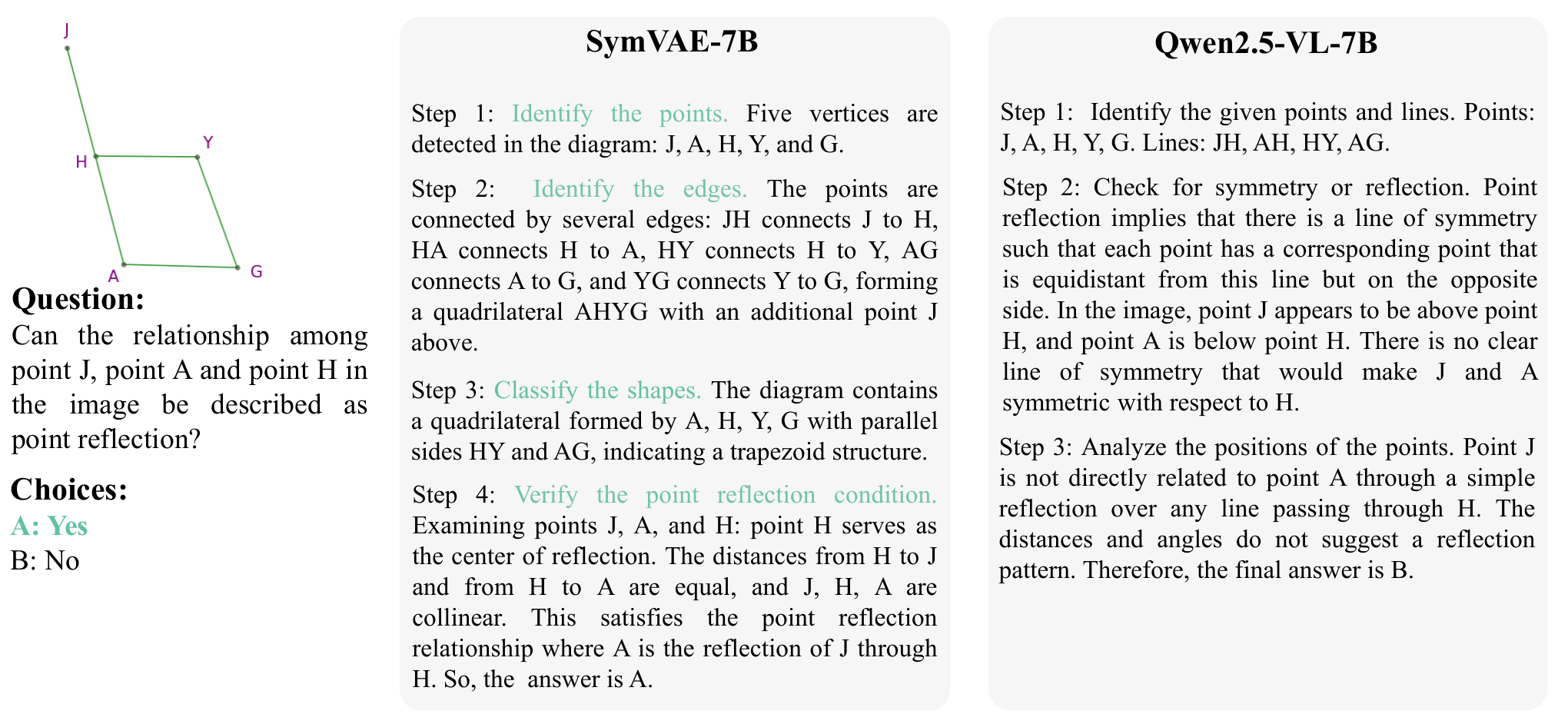}
    \end{subfigure}
    \caption{Emergent cross-lingual translation on MathGlance evaluation samples. The translated chain-of-thoughts are not simple mirrors of the priors embedded in the base model (Qwen2.5-VL-7B); instead, they follow the structured reasoning induced by our logic-form representation.}
    \label{fig:cot_comparison2}
\end{figure}

As shown in Fig.~\ref{fig:cot_demo1}, the model’s reasoning faithfully follows the hierarchical structure encoded in the logic forms: beginning with vertices, then edges, and finally shape attributes before identifying the correct shape category. We hypothesize that this emergent behavior is enabled by the strong NLP reasoning priors of the base model (Qwen2.5-VL-7B). To determine whether this phenomenon reflects genuine cross-lingual transfer or merely mirrors the base model’s priors, we compare our model’s reasoning traces with those generated by the original Qwen2.5-VL-7B. Indeed, the base model exhibits a markedly different reasoning pattern—often starting from global shape descriptions and only later mentioning sides or angles—deviating from the hierarchical primitive structure defined in our logic rules. Similar patterns appear in Figs.~\ref{fig:cot_demo2}-\ref{fig:cot_comparison2}: our model consistently follows the correct reasoning trajectory and produces the correct answer, whereas Qwen2.5-VL-7B generates disorganized or inconsistent reasoning steps that lead to incorrect predictions.

\begin{figure}[t]
    \centering
    \begin{subfigure}{.8\textwidth}
        \centering
        \includegraphics[width=\textwidth]{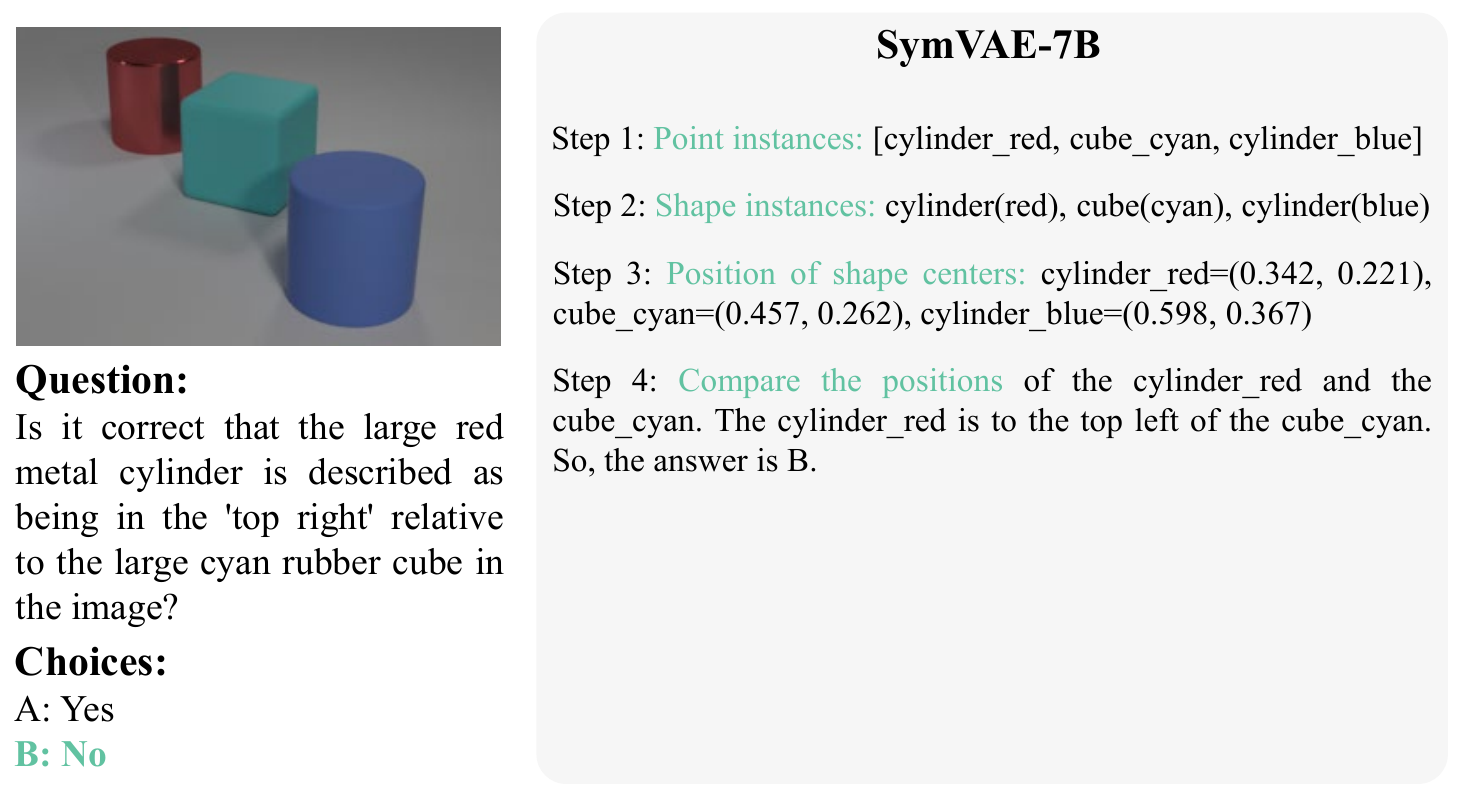}
        \label{fig:cot_demo4}
    \end{subfigure}
    \caption{Emergent cross-modal translation on MathGlance evaluation samples. The model can translate planar geometric vertex points into 3D object attributes.}
    \label{fig:cot_comparison3}
\end{figure}
\begin{figure}[t]
    \centering
    \begin{subfigure}{.8\textwidth}
        \centering
        \includegraphics[width=\textwidth]{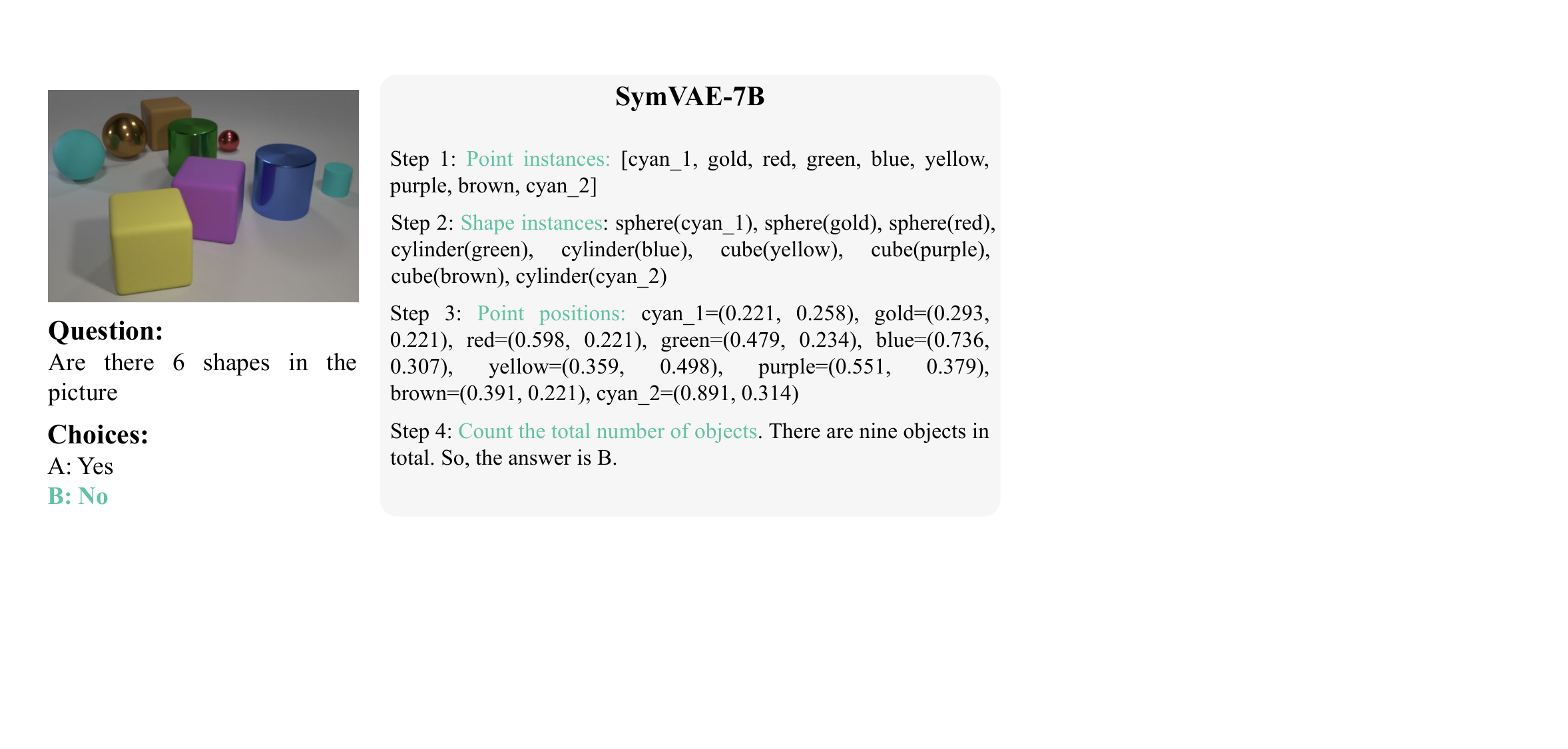}
        \caption{}
        \label{fig:cot_demo5}
    \end{subfigure}
    \hfill
    \begin{subfigure}{.8\textwidth}
        \centering
        \includegraphics[width=\textwidth]{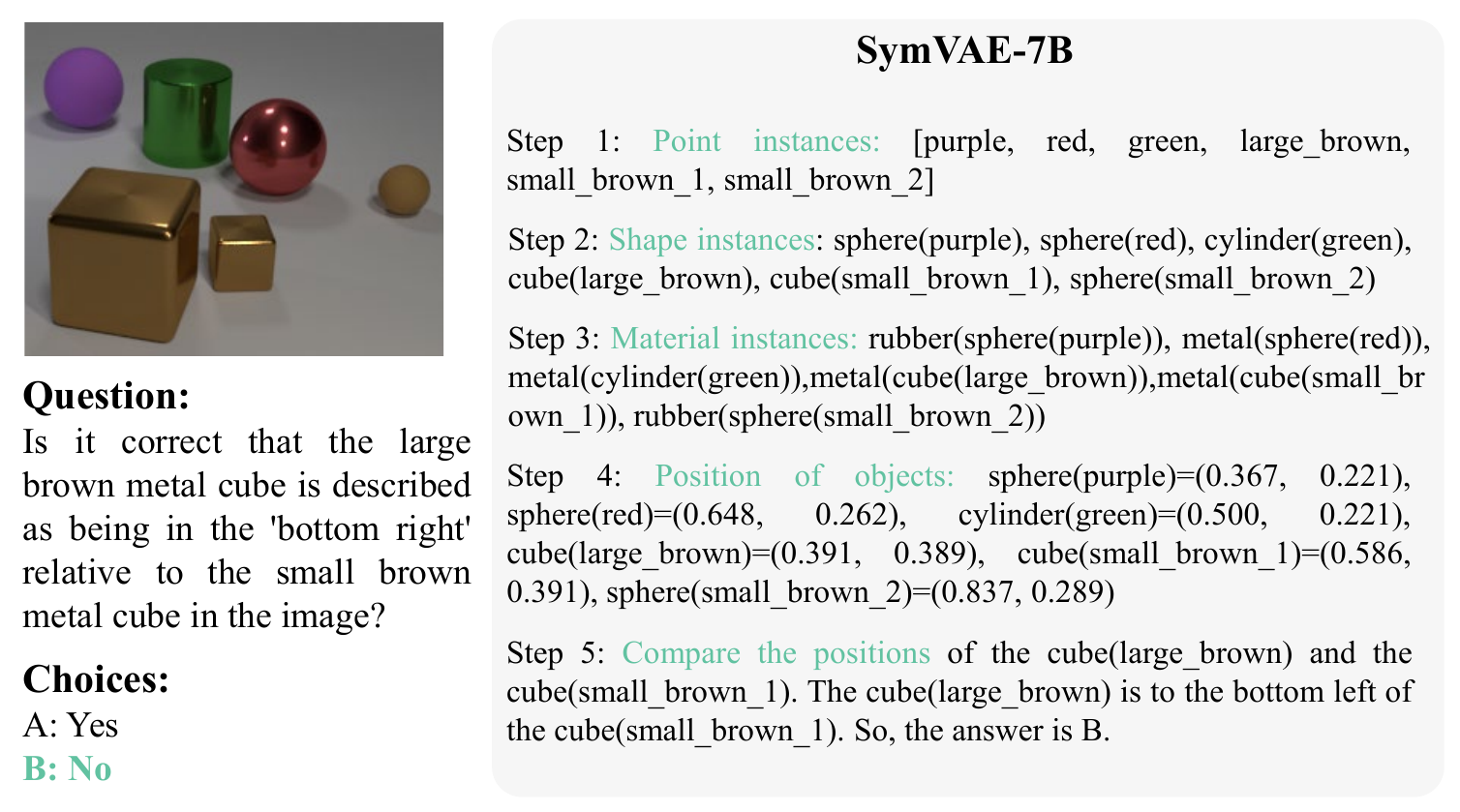}
        \caption{}
        \label{fig:cot_demo6}
    \end{subfigure}
    \caption{Emergent cross-modal translation on MathGlance evaluation samples. The model can translate planar geometric vertex points into 3D object attributes.}
    \label{fig:cot_comparison4}
\end{figure}

\noindent\textbf{Cross-Modal Translation.}
Another interesting finding is an emergent form of cross-modal translation. Although our training data contain only planar geometric diagrams, where the fundamental primitives are points corresponding to 2D vertices, we observe that the model can automatically adapt these primitives when inferring over solid 3D scenes. In particular, the model enriches the point instances with 3D attributes such as object shape, color, and material.  As shown in Fig.~\ref{fig:cot_comparison3}, the inferred point primitives become \texttt{cylinder\_red}, \texttt{cube\_cyan}, and \texttt{cylinder\_blue}, demonstrating a self-adaptation from planar vertices to 3D object descriptors. Furthermore, as illustrated in Figs.~\ref{fig:cot_demo5}--\ref{fig:cot_demo6}, when multiple objects share the same color, the model automatically introduces disambiguating markers (\eg, \texttt{cyan\_1}, \texttt{cyan\_2}) to distinguish them. By explicitly enumerating foreground objects with these extended attributes, the model can more reliably support downstream tasks such as object counting and spatial relationship identification.


\section{Case Studies on Reasoning}
\label{supp:reason}
\begin{figure}[h]
    \centering
    \begin{subfigure}{.9\textwidth}
        \centering
        \includegraphics[width=\textwidth]{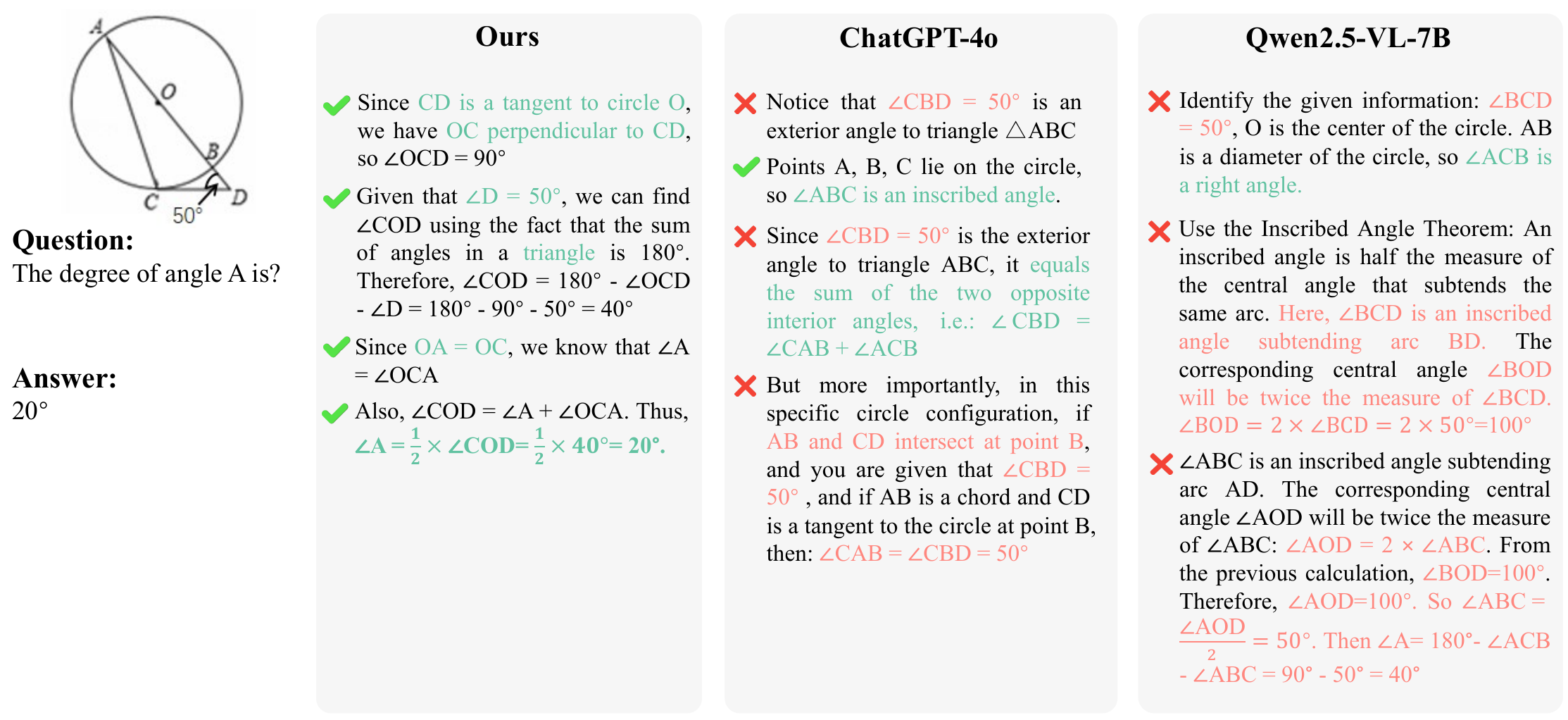}
        \caption{}
        \label{fig:mathverse_demo1}
    \end{subfigure}
    \hfill
    \begin{subfigure}{.9\textwidth}
        \centering
        \includegraphics[width=\textwidth]{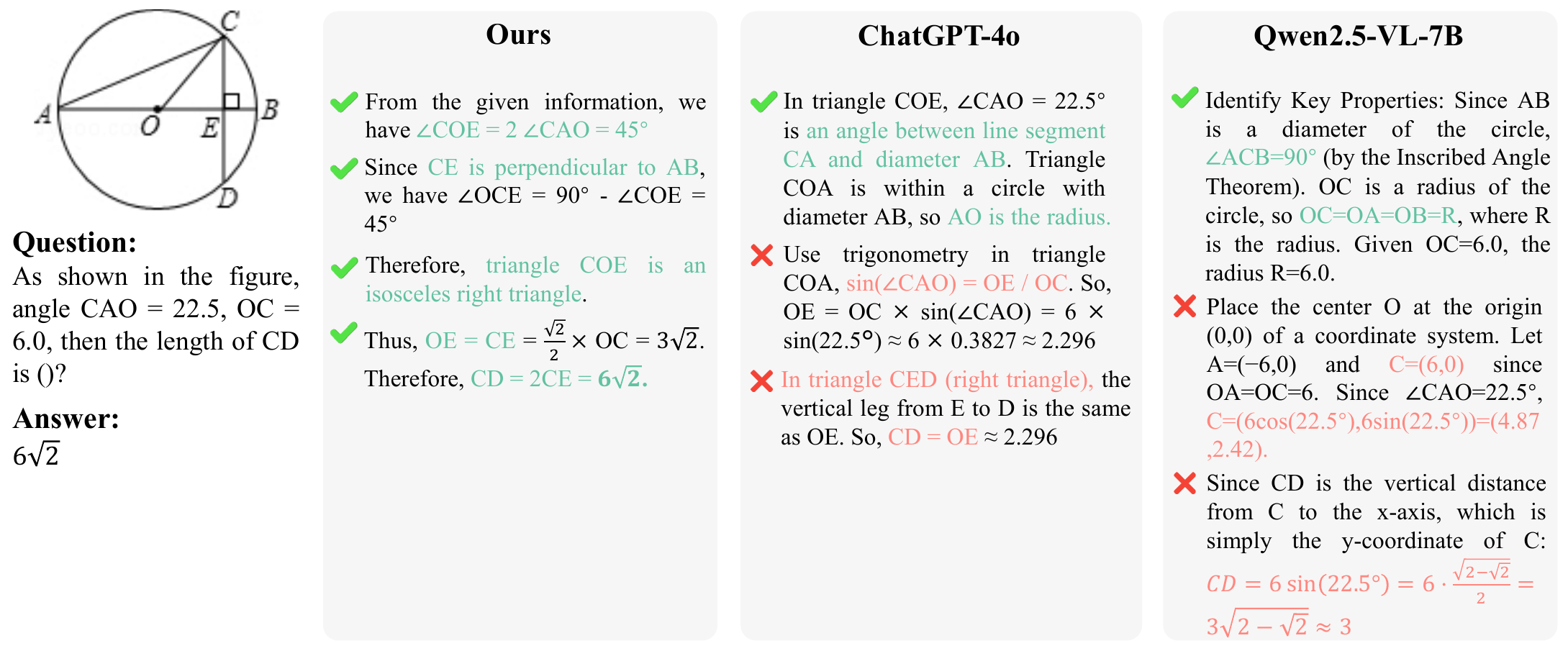}
        \caption{}
        \label{fig:mathverse_demo2}
    \end{subfigure}
    \hfill
    \begin{subfigure}{.9\textwidth}
        \centering
        \includegraphics[width=\textwidth]{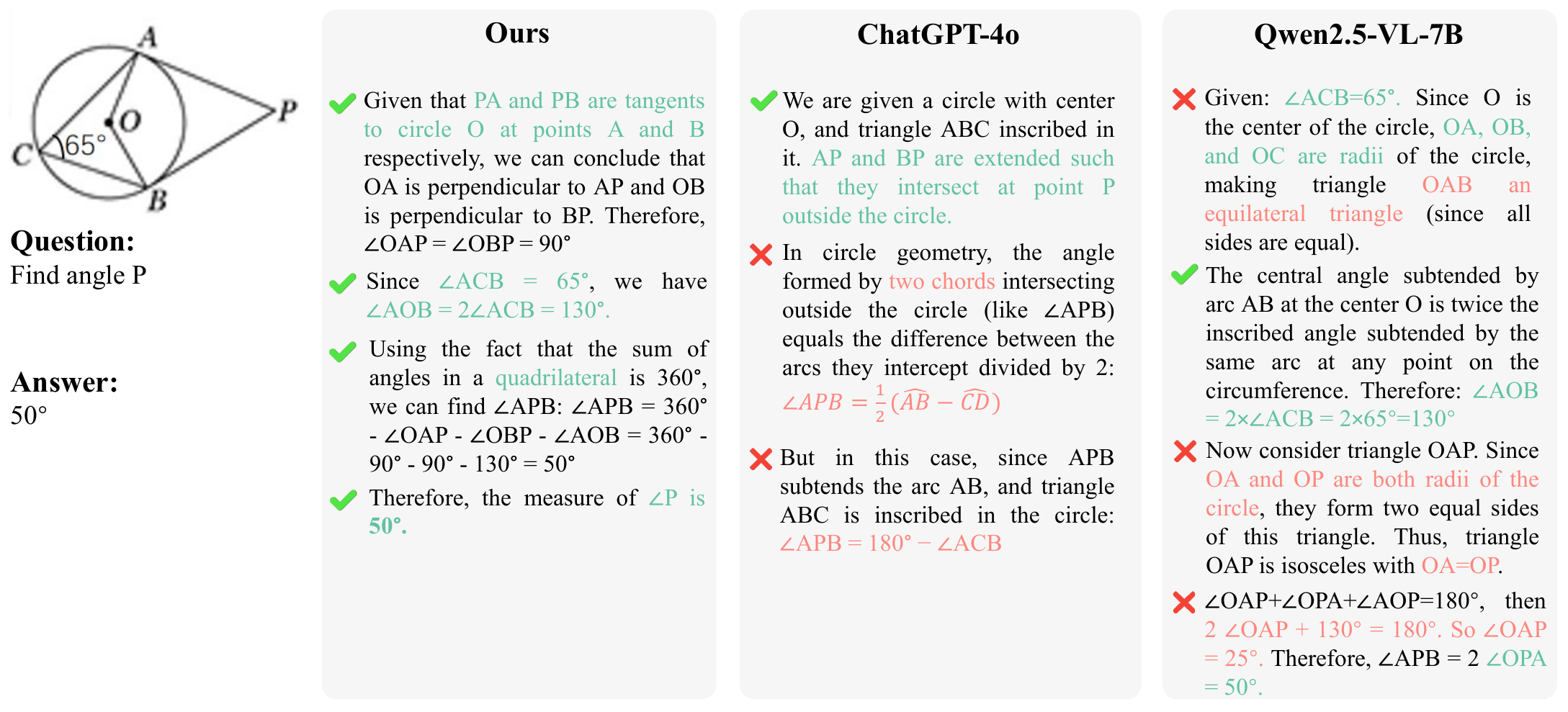}
        \caption{}
        \label{fig:mathverse_demo3}
    \end{subfigure}
    \caption{Response comparisons between \ourvae{}+CoTs-7B, GPT-4o, and Qwen-2.5-VL-7B in MathVerse.}
    \label{fig:response_appendix_1}
\end{figure}

\begin{figure}[h]
    \centering
    \begin{subfigure}{.95\textwidth}
        \centering
        \includegraphics[width=\textwidth]{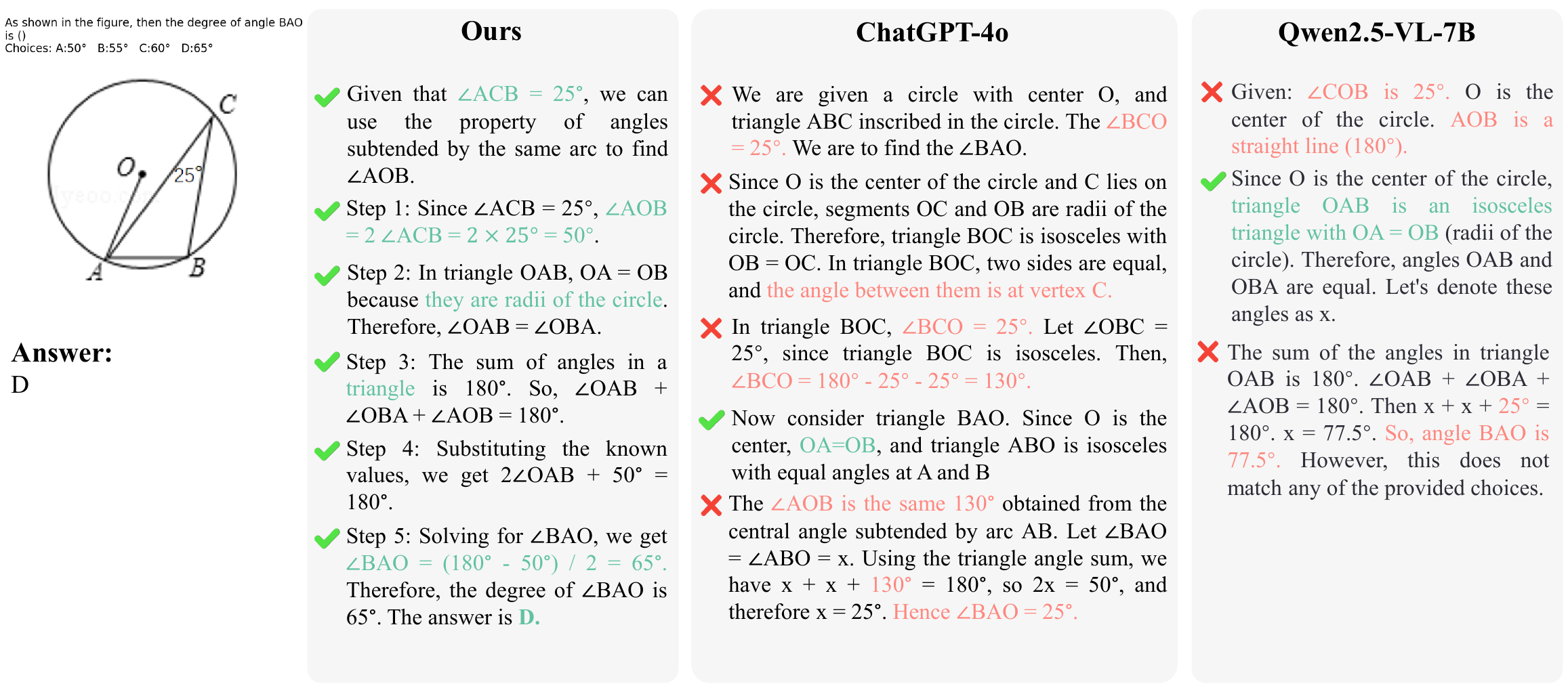}
        \label{fig:mathverse_demo4}
    \end{subfigure}
    \caption{Response comparisons between \ourvae{}+CoTs-7B, GPT-4o, and Qwen-2.5-VL-7B in MathVerse.}
    \label{fig:response_appendix_2}
\end{figure}

Figs.~\ref{fig:response_appendix_1} and \ref{fig:response_appendix_2} present qualitative comparisons on the MathVerse benchmark. Our model consistently produces correct reasoning chains grounded in accurate visual perception, while GPT-4o and Qwen2.5-VL-7B often misidentify geometric configurations, leading to erroneous conclusions. The visual tokens produced by our symbolic vision encoder are more closely aligned with natural-language feature spaces, enabling the model to learn visually grounded reasoning for multimodal problem solving and significantly reducing visual hallucinations.

Specifically, symbolic vision tokens establish a more reliable perceptual foundation for reasoning. Across many examples, the base model fails because it starts with an incorrect perceptual premise (\eg, misidentifying key vertices or an angle), causing the entire reasoning chain to collapse. In contrast, our model corrects these initial perceptual errors, enabling the reasoning process to begin from an accurate visual foundation. For instance, in Fig.~\ref{fig:mathverse_demo1}, the base model incorrectly recognizes the $\angle CBD$, whereas our model correctly identifies it and naturally applies the Angle Sum Theorem to reach the correct answer. Similarly, in Fig.~\ref{fig:mathverse_demo2}, our model accurately perceives the perpendicular relationship, infers the right angle, and proceeds with the appropriate reasoning steps. Beyond accuracy, improved perception enables the model to generate more direct reasoning trajectories rather than overthinking or searching for unnecessarily complex alternatives. By removing perceptual ambiguity, the model can commit earlier to the correct strategy. In the more challenging examples shown in Fig.~\ref{fig:response_appendix_2}, our model demonstrates significantly more stable multi-step reasoning because the intermediate perceptual relations—especially in vision-only settings on MathVerse test samples—are now correctly grounded.


\section{More results}
\label{supp:expresults}
\subsection{Neuro-Symbolic Reasoning with Adaptive Visual-Textual Rewards}

Instead of integrating visual tokens directly into MLLMs—the conventional design of multimodal large language models—we adopt a neuro-symbolic architecture. Specifically, our \ourparser{} generates executable logic forms from diagrams, which are subsequently processed by a mathematical reasoning model (Qwen2.5-Math-7B) to solve multimodal geometric problems. We evaluate this neuro-symbolic pipeline on the GeoQA benchmark.

Concretely, we first apply \ourparser{} to 5K diagrams from Geo170K to generate paired logic forms, which are prepended to the corresponding questions with reasoning chains-of-thought (CoTs). The resulting data serve as supervised fine-tuning (SFT) samples for Qwen2.5-Math-7B. We then apply reinforcement learning (GRPO) using conventional textual answer rewards. Since our logic forms are executable, we further introduce a \textbf{visual reconstruction reward} $r_{\text{vis}}$ (Eq.~\ref{eq:vis}), where reconstructed diagrams are generated from logic forms using our decoder. The total reward thus combines textual accuracy (final answer and format) with visual fidelity, jointly guiding the policy model toward grounded multimodal reasoning.

During training, we observe that rollouts with low visual rewards often produce irrelevant or malformed outputs, including nonsensical or ``garbage'' tokens unrelated to the question. In contrast, samples with higher visual rewards demonstrate coherent reasoning that effectively references visual cues. To mitigate the impact of low-quality visual grounding, we introduce an \textbf{adaptive reward weighting mechanism} that dynamically adjusts the contribution of visual rewards based on mathematical correctness. The key insight is that visual understanding should primarily support correct reasoning. When the model generates incorrect answers ($r_{\text{math}} \leq \tau$), we exponentially decay the visual reward to prevent spurious visual–textual correlations. Conversely, when the answers are correct ($r_{\text{math}} > \tau$), the full visual reward is preserved to reinforce accurate visual grounding.

Formally, the adaptive weight $\gamma(r_{\text{math}})$ is defined as:
\begin{equation}
\gamma(r_{\text{math}}) =
\begin{cases}
1.0, & \text{if } r_{\text{math}} > \tau, \\
\exp(r_{\text{math}} - \tau), & \text{if } r_{\text{math}} \leq \tau.
\end{cases}
\label{supp:eq:vis}
\end{equation}

The modulated visual reward and overall reward are then defined as:
\begin{align}
r_{\text{vis}} &= \alpha \cdot \gamma(r_{\text{math}}) \cdot r_{\text{vis}}, \\
r_{\text{total}} &= w_1 \cdot r_{\text{vis}} + w_2 \cdot r_{\text{math}},
\end{align}
where $r_{\text{vis}}$ denotes the reconstruction-based visual reward and $r_{\text{math}}$ the mathematical correctness reward. We use $\alpha = 1.0$, $w_1 = 0.3$, and $w_2 = 0.7$ by default. This design ensures that visual grounding contributes positively only when it leads to correct mathematical outcomes. The exponential decay for incorrect outputs ($\gamma < 1$) provides smooth gradient feedback and prevents noisy rollouts from dominating policy updates.

Compared to directly integrating visual tokens into MLLMs, our neuro-symbolic system provides explicit interpretability and modular supervision but exhibits limited flexibility—particularly in aligning visual symbolic representations with pretrained reasoning states. We hypothesize that reinforcement learning primarily refines existing reasoning policies rather than inducing new multimodal knowledge, a finding consistent with prior observations~\cite{yue2025does}. We advocate future research toward neuro-symbolic architectures and adaptive, reasoning-aligned visual supervision as a promising direction for developing grounded multimodal intelligence.

\subsection{Close-up results on MathVerse}
Tab.~\ref{supptab:mathverse_results} provides detailed breakdowns of our model's performance on the MathVerse \texttt{test-mini} benchmark, expanding upon the aggregate results presented in the main paper (Sec.~\ref{exp:main_results}).

\section{Limitations}
\label{supp:limitation}

Although our symbolic–visual framework provides clear interpretability and verifiable reasoning, it also introduces several limitations. First, our logic rules are manually designed around planar geometric primitives (points, lines, circles, and polygons). This rule-based design limits flexibility: extending the system to new geometric concepts requires manual modification of the formal grammar and the rendering engine, creating a knowledge-acquisition bottleneck characteristic of classical symbolic AI. In contrast to end-to-end neural approaches that can implicitly learn novel visual patterns through gradient-based adaptation, our symbolic representations cannot autonomously generalize beyond their predefined ontology.

Second, our training data consists exclusively of synthetic planar geometry diagrams with clean and well-structured primitives. Although we observe encouraging cross-modal transfer to 3D object scenes, such generalization remains limited. The system struggles with: (1) hand-drawn diagrams with irregular strokes or occlusions, and (2) real-world images containing cluttered backgrounds and complex topologies. Moreover, the symbolic representation requires precise and complete detection of all primitives. Any missing vertex, line, or relation produces an incomplete logic form that can propagate errors throughout the entire chain, leading to compounding failures.

Finally, for our neuro-symbolic reasoning design, the reasoning performance is still tied to the expressivity of the symbolic representation. When the symbolic abstraction fails to encode a necessary relation, the model cannot recover it through language alone. Thus, improving perception and symbolic completeness remains essential for fully reliable multi-step reasoning.

\noindent\textbf{Future Directions.} 
Rather than relying on hand-crafted logic rules, future work could explore meta-learning approaches that automatically discover primitive vocabularies from data, similar to neural program synthesis. Learning symbolic vocabularies directly from large-scale multimodal corpora would allow the system to expand beyond fixed rule sets while preserving interpretability. Another promising direction is to integrate neural symbolic parsing with differentiable program induction, enabling the symbolic space to grow adaptively for new domains such as chemistry, physics diagrams, or engineering drafts. In addition, incorporating uncertainty estimation into symbolic extraction could improve robustness and reduce cascading perceptual errors during multi-step reasoning. Finally, neuro-symbolic cycles, where a symbolic encoder and a neural reasoner interact in a bidirectional loop, may yield systems that combine flexible neural pattern recognition with rigorous symbolic verification.

\clearpage

\clearpage

\begin{table*}[!t]
\centering
\caption{Performance comparison on MathVerse testmini set. We evaluate models across different visual dependency levels: \textit{Text Dominant} (text-heavy problems), \textit{Text Lite} (moderate text), \textit{Vision Intensive} (requires visual understanding), \textit{Vision Dominant} (primarily visual), and \textit{Vision Only} (purely visual reasoning). Results show accuracy (\%) for each category.}
\vspace{-.5em}
\label{supptab:mathverse_results}
\renewcommand\arraystretch{1}
\setlength{\tabcolsep}{3mm}{
\resizebox{0.9\linewidth}{!}{
\begin{tabular}{l|c|c|c|c|c|c|c}
\hline
\textbf{Model} & \textbf{Base LLM} & \textbf{All} & \textbf{Text Dom.} & \textbf{Text Lite} & \textbf{Vision Int.} & \textbf{Vision Dom.} & \textbf{Vision Only} \\
\hline
\rowcolor{green!15}\multicolumn{8}{l}{\textit{Baselines}}   \\ \hline
Random Chance & - & 12.4 & 12.4 & 12.4 & 12.4 & 12.4 & 12.4 \\
Human & - & 67.7 & 71.2 & 70.9 & 61.4 & 68.3 & 66.7 \\
\hline
\rowcolor{green!15}\multicolumn{8}{l}{\textit{Large Language Models}}   \\ \hline
ChatGPT & - & 26.1 & 33.3 & 18.9 & - & - & - \\
GPT-4 & - & 33.6 & 46.5 & 46.5 & - & - & - \\
\hline
\rowcolor{green!15}\multicolumn{8}{l}{\textit{Closed-Source Multimodal LLMs}}   \\ \hline
Qwen-VL-Plus~\cite{bai2023qwenvl} & - & 11.8 & 15.7 & 11.1 & 9.0 & 13.0 & 10.0 \\
Qwen-VL-Max~\cite{bai2023qwenvl} & - & 25.3 & 30.7 & 26.1 & 24.1 & 24.1 & 21.4 \\
\hline
\rowcolor{green!15}\multicolumn{8}{l}{\textit{Open-Source Multimodal LLMs}}   \\ \hline
LLaMA-Adapter V2~\cite{gao2023llamaadapterv2}& LLaMA-7B & 5.7 & 6.2 & 5.9 & 6.1 & 4.2 & 6.1 \\
ImageBind-LLM~\cite{han2023imagebindllm}  & LLaMA-7B & 9.2 & 11.4 & 11.3 & 8.9 & 11.2 & 3.4 \\
mPLUG-Owl2~\cite{ye2024mplugowl2} & LLaMA-7B & 5.9 & 6.6 & 6.3 & 6.3 & 5.6 & 4.9 \\
SPHINX-Plus ~\cite{gao2023sphinx}& LLaMA2-13B & 12.2 & 13.9 & 11.6 & 11.6 & 13.5 & 10.4 \\
SPHINX-MoE~\cite{gao2024sphinx} & Mixtral-8×7B & 15.0 & 22.2 & 16.4 & 14.8 & 12.6 & 9.1 \\
G-LLaVA~\cite{gao2023gllava} & LLaMA2-7B & 16.6 & 20.9 & 20.7 & 17.2 & 14.6 & 9.4 \\
LLaVA-1.5~\cite{liu2024visual} & Vicuna-13B & 7.6 & 8.8 & 7.6 & 7.4 & 7.4 & 6.9 \\
ShareGPT4V~\cite{chen2024sharegpt4v}& Vicuna-13B & 13.1 & 16.2 & 16.2 & 15.5 & 13.8 & 3.7 \\
Math-LLaVA~\cite{shi2024mathllava} & Vicuna-13B & 19.0 & 21.2 & 19.8 & 20.2 & 17.6 & 16.4 \\
LLaVA-NeXT~\cite{liu2024llavanext} & LLaMA3-8B & 19.3 & 24.9 & 20.9 & 20.8 & 16.1 & 13.8 \\
Qwen2.5-VL-7B~\cite{bai2023qwenvl} & Qwen2.5-7B & 49.2 & 58.4 & 52.7 & 33.2 & 30.6 & 21.1 \\
\hline
\rowcolor{green!15}\multicolumn{8}{l}{\textit{Our Models}}   \\ \hline
\bf \ourvae{}+CoTs-7B & Qwen2.5-7B & \bf 51.8 & \bf 61.3 & \bf 55.2 & \bf 35.2 & \bf 33.4 & \bf 24.9 \\
\hline
\end{tabular}}}
\vspace{-0.3cm}
\end{table*}

\end{document}